\documentclass{article}

\usepackage{microtype}
\usepackage{graphicx}
\usepackage{wrapfig}
\usepackage{subfigure}
\usepackage{booktabs} 
\usepackage[normalem]{ulem}

\usepackage{hyperref}

\usepackage[accepted]{icml2024}

\usepackage{amsmath}
\usepackage{amssymb}
\usepackage{mathtools}
\usepackage{amsthm}

\usepackage[capitalize,noabbrev]{cleveref}

\theoremstyle{plain}

\theoremstyle{definition}

\theoremstyle{remark}

\usepackage[textsize=tiny]{todonotes}

\newcommand{\haggai}[1]{%
\textcolor{green}{[HM: \textit{#1}] }%
}

\newcommand{\ef}[1]{%
\textcolor{orange}{[EF: \textit{#1} ]}%
}

\newcommand{\revision}[1]{\textcolor{black}{#1}}

\icmltitlerunning{Improved Generalization of Weight Space Networks via Augmentations}

\begin{document}

\twocolumn[
\icmltitle{Improved Generalization of Weight Space Networks via Augmentations}




\begin{icmlauthorlist}
\icmlauthor{Aviv Shamsian}{biu}
\icmlauthor{Aviv Navon}{biu}
\icmlauthor{David W. Zhang}{amst}
\icmlauthor{Yan Zhang}{sam} \\~\\
\icmlauthor{Ethan Fetaya}{biu}
\icmlauthor{Gal Chechik}{biu,comp}
\icmlauthor{Haggai Maron}{tech,comp}
\end{icmlauthorlist}

\icmlaffiliation{biu}{Bar-Ilan University}
\icmlaffiliation{amst}{University of Amsterdam}
\icmlaffiliation{comp}{NVIDIA Research}
\icmlaffiliation{tech}{Technion}
\icmlaffiliation{sam}{Samsung\,-\,SAIT AI Lab, Montreal}

\icmlcorrespondingauthor{Aviv Shamsian}{aviv.shamsian@biu.ac.il}
\icmlcorrespondingauthor{Aviv Navon}{aviv.navon@biu.ac.il}

\icmlkeywords{Machine Learning, ICML}

\vskip 0.3in
]



\printAffiliationsAndNotice{} 

\begin{abstract}

Learning in deep weight spaces (DWS), where neural networks process the weights of other neural networks, is an emerging research direction, with applications to 2D and 3D neural fields (INRs, NeRFs), as well as making inferences about other types of neural networks. Unfortunately, weight space models tend to suffer from substantial overfitting. We empirically analyze the reasons for this overfitting and find that a key reason is the lack of diversity in DWS datasets.  While a given object can be represented by many different weight configurations, typical INR training sets fail to capture variability across INRs that represent the same object. To address this, we explore strategies for data augmentation in weight spaces and propose a MixUp method adapted for weight spaces. We demonstrate the effectiveness of these methods in two setups. In classification, they improve performance similarly to having up to 10 times more data. In self-supervised contrastive learning, they yield substantial 5-10\% gains in downstream classification.

\end{abstract}

\section{Introduction}
\label{intro}

\begin{figure*}[t]
\centering
    \begin{subfigure}{
    \includegraphics[width=0.4\linewidth]{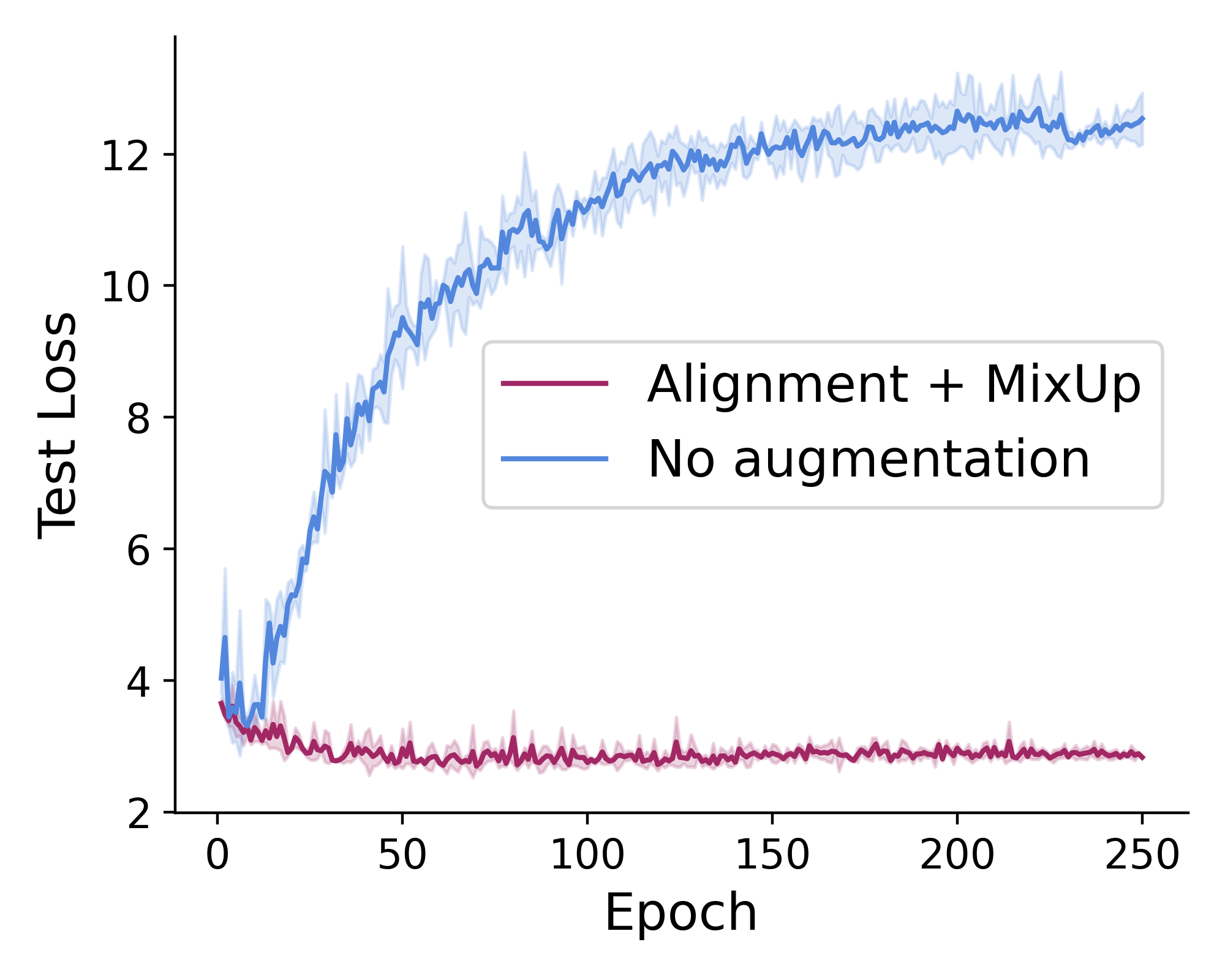}
    }
    \end{subfigure}
    \begin{subfigure}{
    \includegraphics[width=0.4\linewidth]{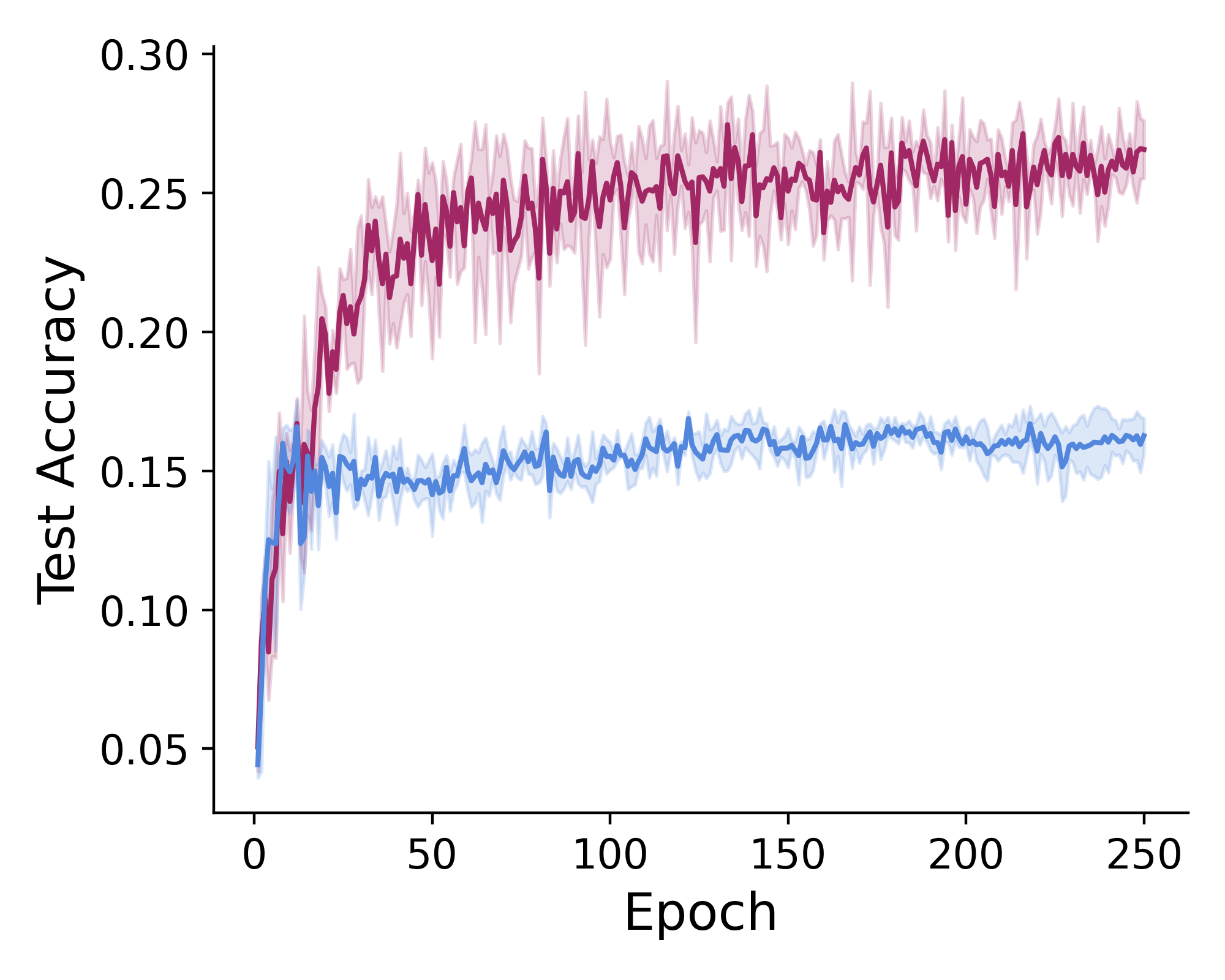}
    }
     \end{subfigure}
    \caption{\textit{Effect of our Alignment + MixUp augmentation on ModelNet40}: MixUp without augmentation (blue) shows major signs of overfitting from the increasing test loss, while MixUp + alignment (maroon) mitigates this overfitting and improves the DWS's accuracy.}
\label{fig:MixUp_effect}
\end{figure*}

Learning in deep weight spaces (DWS) is the task of training models that take the weights of other deep neural networks as input 
\citep{eilertsen2020classifying, unterthiner2020predicting, andreis2023set}. 
It provides a way to infer properties of \emph{neural networks} themselves, for tasks like ranking models by their predicted performance without applying them to a test set.
Deep weight spaces have a complex geometrical structure thanks to the various symmetries of the weights and biases in deep learning models \cite{hecht1990algebraic}.
Architectures for weight spaces benefit significantly from taking this complex structure into account
\citep{navon23dws,zhou2023permutation,zhou2023neural,zhang2023neural,lim2023graph}. 
However, these methods often run into overfitting and generalization issues. Standard techniques to reduce overfitting are difficult to apply due to the complex structure, or failing to improve results entirely. Alleviating this problem is the core goal of this paper.



\begin{figure}[t]
    \centering
\includegraphics[width=0.95\linewidth]{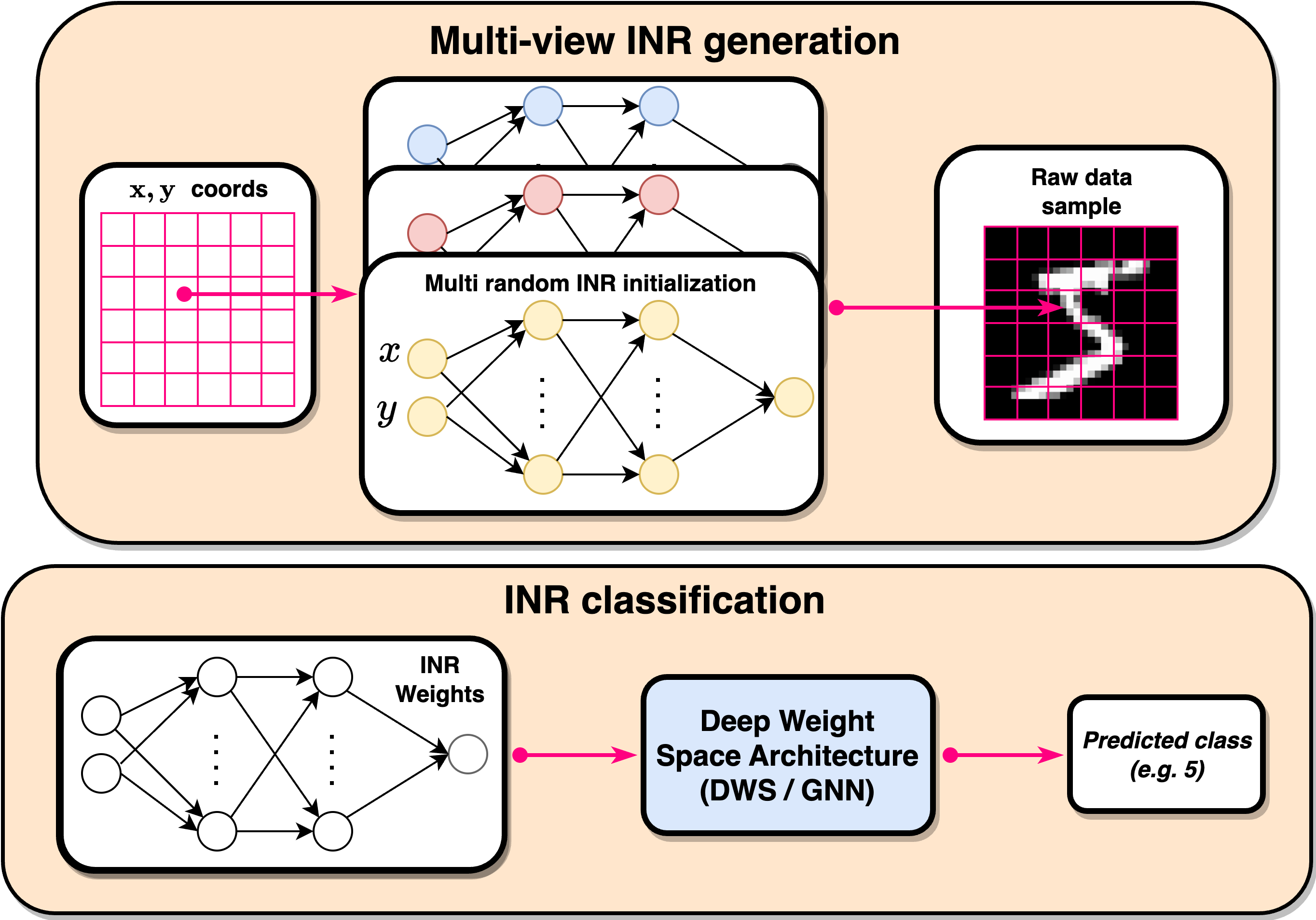}
    \caption{\textit{Illustration of multiview INR generation and INR classification task.} The INR is trained to receive $x,y$ coordinates and map them to the corresponding grayscale value in the original raw data (top panel). The trained INR is fed into the weight space architecture to perform classification (bottom panel).}
    \label{fig:overview}
\end{figure}

To illustrate these generalization difficulties, consider the the problem of using a deep model to process an  
Implicit Neural Representation (INR) \cite{mescheder2019occupancy,park2019deepsdf, mildenhall2021nerf} --  a neural network that represents an image or a 3D shape (\cref{fig:overview}). Training a classifier over 
the weights of INRs performs far worse than 
training standard CNNs or MLPs on the original raw data. As a concrete example, current state-of-the-art for 3D shape classification by processing weights of INRs achieves only 16\% accuracy on ModelNet40 \cite{wu20153d} (see \cref{fig:MixUp_effect}), compared to 90\% achieved by applying neural networks directly to point cloud representation of the same shapes \cite{atzmon2018point,wang2019dynamic}.  The reasons for this performance gap are still not well understood.


Here, we study the causes for that  generalization gap and propose ways to mitigate them. 
%
We argue that typical training workflows in DWS 
fail to represent the variability across different weight representations of the same object well.  
In particular,  a single object can be represented by a huge 
number of different weight space representations, which we call neural views, but DWS datasets often only use a single view for each object. This lack of cross-view diversity confounds two different generalization problems that DWS learning needs to address: \textit{generalization to new neural views} and \textit{generalization to new objects}. 


To address this issue, we first 
empirically study the effects of neural views on generalization to new objects and gain a key insight (\cref{sec:empirical}): \textbf{training with multiple neural views 
improves generalization to unseen objects}. 
At first look, it appears that this insight can be leveraged by augmenting any object with many neural views to represent it, for example by varying weight initialization. Unfortunately, this na\"ive approach is not practical since it involves training a large number of deep models for each data object.

To increase the effective size of the input set without training additional neural views, we propose new data augmentation techniques tailored for DWS. These techniques  transform any given input (a combination of weights and biases) 
on-the-fly to increase diversity while \emph{preserving} the functions represented by these input weight vectors. Data augmentation is widely applied in domains like images and text, but it is not known yet what augmentation strategies would be effective for learning in deep weight spaces.


To answer this question, we first categorize known augmentation techniques into two groups (\cref{sec: weight_space_aug}): (i) input space augmentations which reflect transformations of the input object, such as rotating a 3D object; (ii) data-agnostic augmentations such as adding noise or masking. Next, we propose a third, novel family of weight space-specific augmentations that leverages architectural symmetries in neural architectures, such as symmetries in activation functions. 

Then, we present a new data augmentation scheme based on the MixUp approach \cite{zhang2017mixup}  for weight spaces (\cref{sec:MixUp}). Unlike MixUp for dense vectors and images, applying MixUp directly to weight space elements is not straightforward due to the permutation symmetries of weight spaces. Specifically, the weights of two independently trained models are rarely aligned, therefore directly interpolating them may not yield a meaningful model \citep{ainsworth2022git}.  
We address this difficulty and develop several variants of weight space MixUp, building on recent weight space alignment and merging algorithms \citep{ainsworth2022git,pena2023re}. \cref{fig:MixUp_effect} illustrates training with and without our proposed weight space MixUp approach, and shows that weight space MixUp substantially mitigates overfitting.



We conduct extensive experiments on three types of INR datasets: grayscale images (FMNIST), color images (CIFAR10), and 3D shapes (Modelnet40). Our results indicate that data augmentation schemes, and specifically our proposed weight space MixUp variants, can enhance the accuracy of weight space models by up to 18\%, equivalent to using 10 times more training data. Moreover, we show the efficacy of our augmentation schemes in a Self-Supervised Learning (SSL) setup, where using our augmentations in a contrastive learning framework yields substantial performance gains
of 5-10\% in downstream classification. 

Our key contributions are as follows: (1) We investigate the issue of overfitting in deep neural network weight spaces and propose mitigating this problem through weight space data augmentation techniques. (2) We categorize existing weight space augmentation methods and discuss their limitations. (3) We introduce new families of weight space augmentations including a novel weight space mixup approach. (4) We conduct extensive experiments analyzing the impact of different weight space augmentations on model generalization as well as their effectiveness in SSL setups.





\section{Preliminaries}\label{sec:prelim}

Learning in DWS could apply to various architectures. We focus here on the multi-layer perceptron (MLPs) architecture because it is a widely used architecture for INRs. 

\textbf{Objects and views.} In this paper, we distinguish between an original input object, such as an image or a 3D mesh, and its \textit{neural view}, which is a neural representation of that object. Specifically, a neural view $v$ of an object $o$ is the 
set of weights and biases of multilayer perceptrons (MLPs) that implicitly represent that object. Importantly, since neural models are highly over-parametrized and  DWS models do not account for all the symmetries in these models,  there is a one-to-many mapping between a given object and the many valid neural views that represent it. One can conceptualize this mapping by defining a conditional probability distribution over possible views, $p(v|o)$, which is determined by many factors, including the training procedure used for training the INRs, the initialization scheme for training, the hyperparameters selected, and so on. Therefore, it would be important for the learning process to take this distribution into account as well.


 %
\textbf{MLPs.} Following \citet{navon23dws,navon2023equivariant}, we define an $M$-layer MLP architecture using the following equations:
\begin{equation}\label{eq:mlp}
    f(x)= x_M, \quad x_{m+1}=\sigma(W_{m+1} x_m +b_{m+1}), \quad x_0=x,
\end{equation}
where $x_m \in \mathbb{R}^{d_m}$, $W_m \in \mathbb{R}^{d_{m}\times d_{m-1}}$, $b_m \in \mathbb{R}^{d_m}$, and $\sigma$ is a pointwise activation function.
We define the weight space vector $v=[w,b]=[W_m,b_m]_{i=1}^M$ to be the concatenation of weight matrices and bias vectors.
An important property of weight spaces is that they have symmetries, meaning that they have multiple equivalent representations related by a group element \cite{hecht1990algebraic}. Specifically, we notice that for any sequence of permutation matrices $p=(P_1,\dots, P_M)$ (with appropriate dimensions) we can define a new weight space vector $v'=[w',b']=[W'_m,b'_m]_{i=1}^M$ via the following equation following the notation in \citet{navon23dws}:
\begin{align} \label{eq:sym}
    W_1'&= P_1 W_1,  W_{M}' =   W_M P^T_{M-1} ,  
    \text{ and } \notag \\
    ~W_m'&= P_m W_m P^T_{m-1}, \forall m\in [2,M-1] \\
    b_1' &= P_1 b_1,
    b_{M}' = b_M,  \text{ and } \notag\\
    b_m'&= P_{m} b_m, \forall m\in [2,M-1] \notag.
\end{align}
This equation, which we denote concisely by $v'=p\cdot v$ defines a group action on the weight space containing $v,v'$, and a symmetry in this space since networks parameterized by both weight vectors ($v,v'$) represent the same function.

\section{Generalization in Weight Space}
\label{sec:empirical}

A unique aspect of training weight space models in general, and training ones for INRs in particular, is that even after accounting for the permutation symmetries in \cref{eq:sym} by using equivariant architectures, multiple different views can still represent the same raw objects. This can happen when the equivariant architecture fails to account for certain symmetries or when INRs represent objects that are nearly identical but not exactly the same. Here, we design several experiments to answer key questions in weight space generalization which are currently unexplored: Can additional neural views improve generalization? How many are necessary to prevent overfitting, and under what conditions?
%




\begin{figure}[t]
\centering
    \begin{subfigure}[One view]{
    \includegraphics[width=0.46\linewidth]{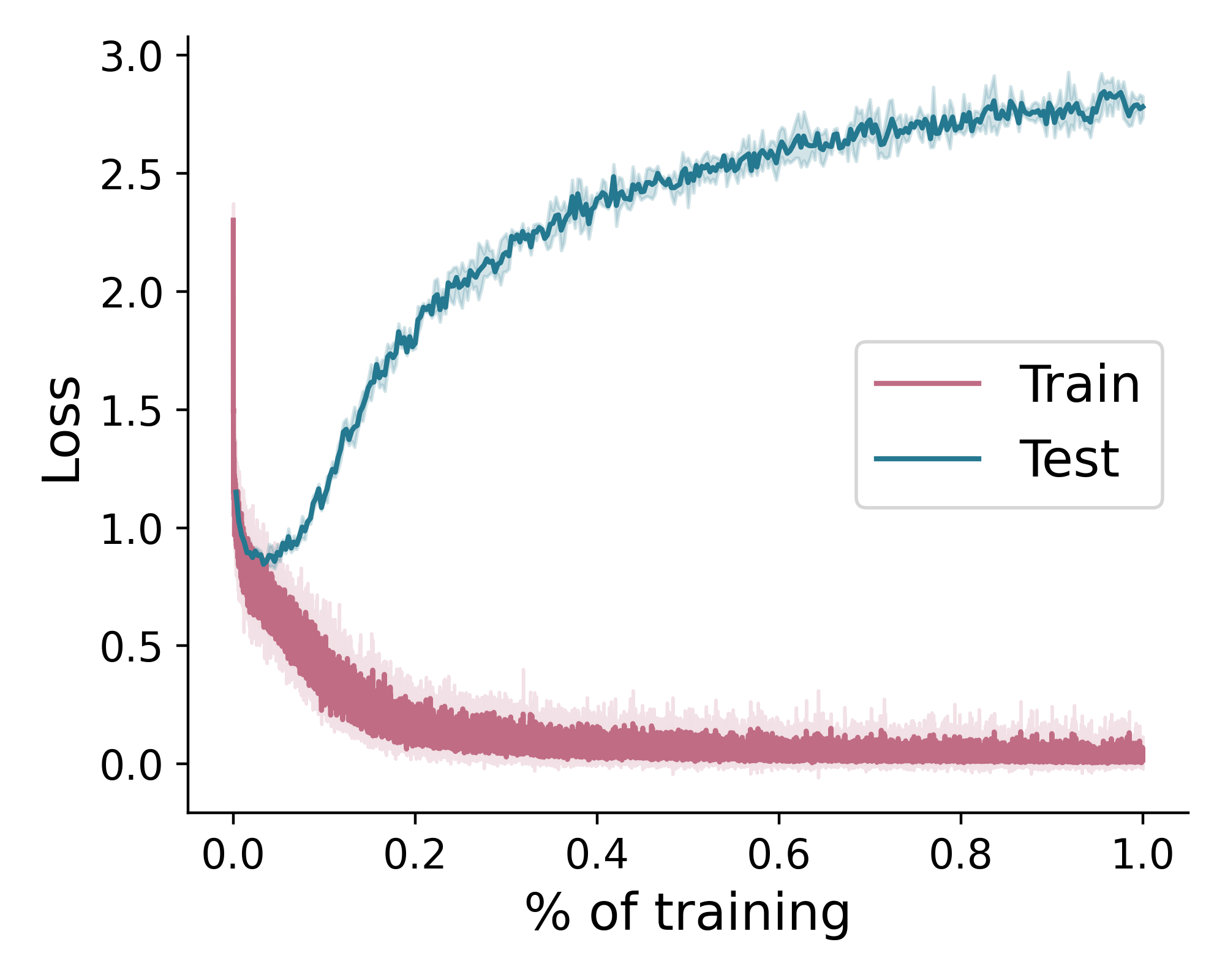}
    }
    \end{subfigure}
    \begin{subfigure}[Ten views]{
    \includegraphics[width=0.46\linewidth]{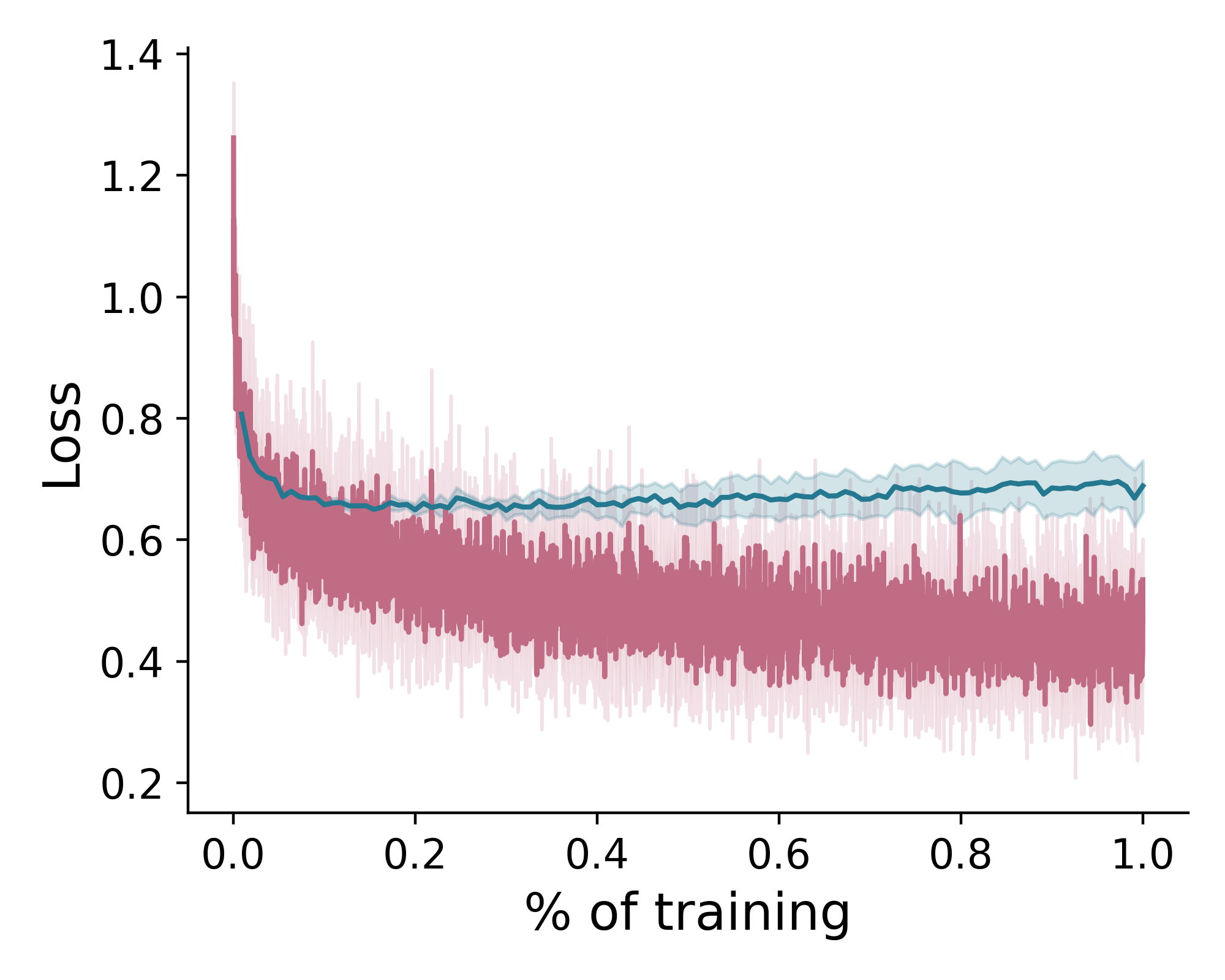}
    }
     \end{subfigure}
    \caption{\textit{Overfitting of weight space architectures on the FMNIST dataset}: We visualize train 
and test 
losses for DWS \citep{navon23dws} on the ModelNet40 dataset with 1 or 10 trained input networks per point cloud (views). Notably, DWS tends to overfit early during training, even when using more data.}
\label{fig:overfit}
\end{figure}

\textbf{Experiment (1): Overfitting of weight-space models.} 

This initial experiment is intended to demonstrate the overfitting problem in DWS, and how the number of views used for training data affects it. To this end, we use two datasets containing either 1 or 10 neural views per object and use the standard train/test split. As shown in Figure \ref{fig:overfit}, models trained on a single view (left panel) start substantially overfitting after only $5-10\%$ of training iterations. In contrast, models trained on $10$ views per object (right panel) suffer from less overfitting and generalize better. The results of this first experiment suggest that DWS models suffer from severe overfitting and that training with multiple views per training image can be used to 
partially mitigate it.


\textbf{Experiment (2): Comparing generalization to new neural views vs.\ new images.}

\begin{figure}[t]
    \centering
\includegraphics[width=0.85\linewidth]{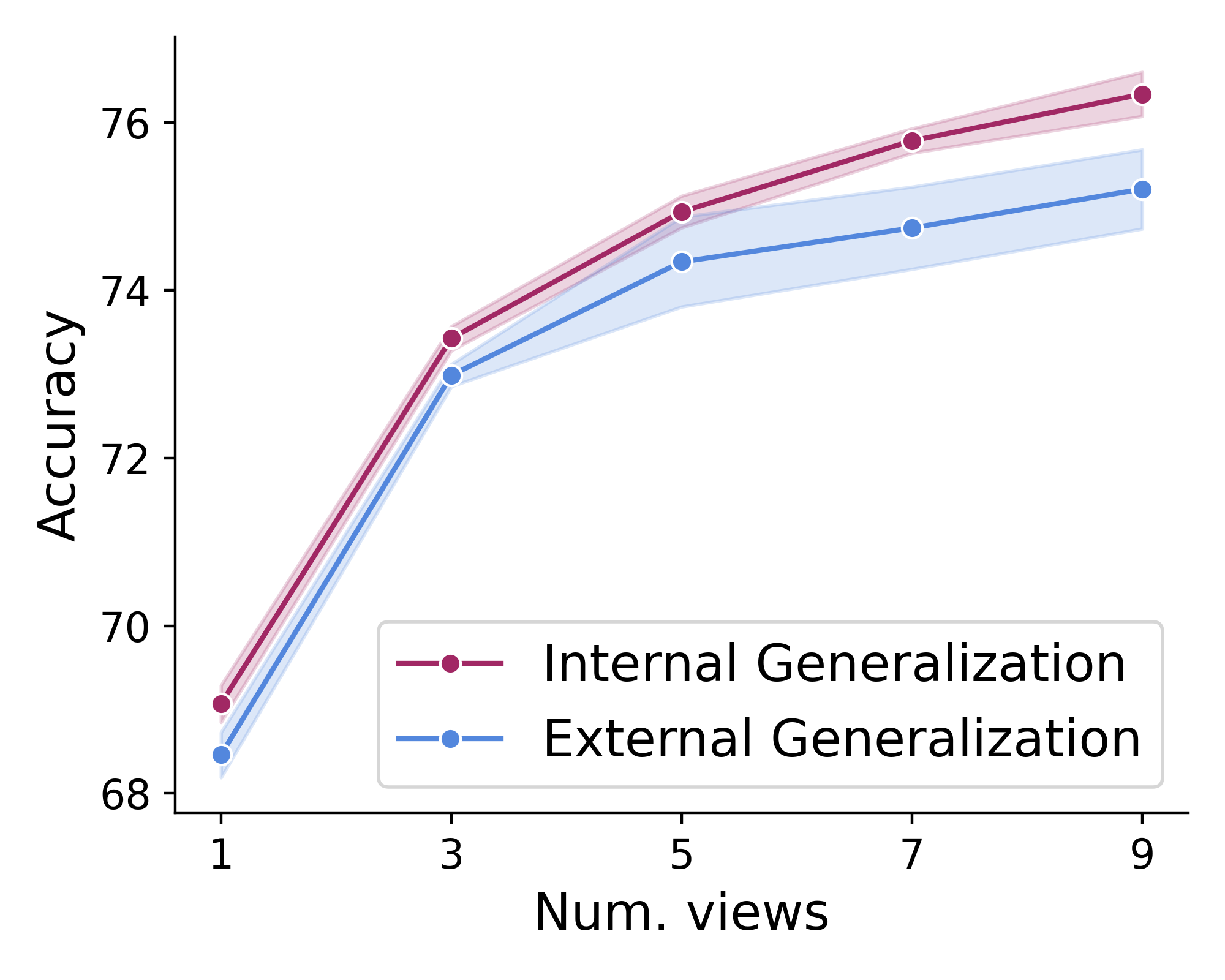}
    \caption{\textit{Internal vs. external generalization}: We visualize DWS performance on internal and external FMNIST splits with varying a number of neural views. There is a relatively small difference between the two types of generalization.}
    \label{fig:gen_types}
\end{figure}

Building on the overfitting experiment, we next evaluate two types of generalization: (1) \emph{internal} -- to unseen views of training images, and (2) \emph{external} -- to views of test images. Models were trained on datasets containing multiple views, ranging from one to nine per training image. We then evaluate the trained models using two test sets: one with new views of training images (internal test set), and one with views of test images (external test set).
As shown in Figure \ref{fig:gen_types}, adding more training views boosts performance on both test sets. While the performance on the internal dataset (purple curve) is slightly better, adding additional training views still significantly improves external generalization (blue curve). Thus, we can deduce that training with multiple views of training images provides valuable information for generalizing to views of test images. 
The experiment also revealed that one view alone is not sufficient to represent a training image -- this can be observed by examining the internal generalization curve (purple), which indicates that additional training views improve the generalization to new views of training images. In line with the expectation, the generalization gain decreases slightly as the number of views increases, but is still significant even when training with 9 views per image. 


\textbf{Experiment (3): What data is more effective for training, additional images or additional neural views?}

Here, we directly compare augmenting the training set with more views versus more objects while keeping the total number of views fixed. Specifically, we generate several datasets with 10K total training INRs and vary the number of views $v$ per object from 1 to 10 which results in training sets with 10K to 1K unique images. 
%
We observe that training with additional views of training images is just as effective in this setup as training with additional views of unseen images. Specifically, our model attains on-par test accuracy when optimized on different data variations (see \cref{fig:view_vs_objects}). 
This implies that while the diversity in training images is important, exposure to additional views can effectively contribute to generalization,  and in certain cases even compensate for the lack of image diversity. 


\textbf{Experiment (4): Is it possible to improve DWS generalization by constraining model capacity?}

One simple approach to prevent overfitting in deep learning is reducing the model's capacity. Here, we evaluate our model with a varying number of trainable parameters and DWS layers. Specifically, we control the number of learnable parameters by changing the hidden dimension and number of hidden layers in the DWS model. The results indicate that such a reduction does not help in this case -- the most expressive model (i.e. DWS with $32$ hidden dim and $4$ hidden layers), although still overfitting, has the best test performance  (see \cref{fig:dws_capacity}). One way to interpret this result is that the overfitting problem stems from a real lack of diversity in training data, rather than from an overly expressive architecture.

\textbf{Discussion.} In summary, our key findings are: (1) adding views of training images significantly boosts generalization to entirely unseen images (first two experiments); (2) multiple views are essential for a faithful representation of images with INRs (second experiment);  (3) using more training views can yield competitive results to using additional unseen training images (third experiment) (4) overfitting cannot be mitigated by simply restricting the model's representation capacity (fourth experiment).
As stated earlier, training additional views can be time-consuming. In the subsequent sections, we explore ways to generate new views on the fly from the training views.

\section{Weight space data augmentation methods}
This section discusses data augmentation schemes for weight spaces, namely methods for generating new views efficiently. We begin by classifying current weight space augmentations and then present weight space mixup schemes.

\subsection{A classification of data augmentation schemes}\label{sec:taxonomy}
Three families of data augmentation schemes are presented here, two of which have been described in previous papers, and one that is entirely new.

\textbf{Input space augmentations.}
When using INRs as inputs, we make a distinction between the input data for our model (i.e., the weight space inputs) and the original data that the INRs represent (e.g., images). Input space augmentations are transformations that can be applied to the weights of an INR to reflect a transformation applied to the original data. Specifically, many conventional data augmentation schemes, like random rotations, translations, and scalings can be utilized when learning with INRs. As demonstrated in \citet{navon23dws}, these augmentations can often be applied to INRs by performing the relevant geometric transformations on the INR input coordinates. For example, rotating the object represented by an INR by a random rotation R can be accomplished by replacing the first weight matrix $W_1$ of the INR with $W_1R$. While this technique generates new views of objects and aligns with previous data augmentation methods, one limitation is that it only affects the first layer weights.


\textbf{Data-agnostic augmentations.} Data-agnostic augmentation techniques are applicable to any data modality. \revision{\citet{schurholt2021self} proposed several data-agnostic augmentations which include familiar methods such as} (1) masking, which randomly nullifies a fraction of input weights during training, and can be seen as an application of dropout~\cite{srivastava2014dropout} to the input weights; (2) threshold-based masking, which removes input weights with magnitudes below a defined threshold; (3) random Gaussian noise addition to the input weights. While these schemes work well for many data modalities, their effectiveness might be limited compared to weight space-specific augmentations since they do not take the structure of weight spaces into account. \revision{It is important to note that the input weights and biases could be augmented by applying random permutations according to Equation \ref{eq:sym} \cite{langosco2023detecting, schurholt2022hyper}. However, this paper does not explore such augmentations since we employ permutation invariant architectures  \cite{navon2023equivariant, zhang2023neural}.}

Last but not least, we present a novel family of augmentations based on the structure of neural networks. 

\textbf{Neural network specific augmentations.}
\label{sec: weight_space_aug}
These augmentations refer to methods that rely on the exact structure or components of the neural network that the weight space input comes from. 
A major source of additional symmetries we make use of in this paper stems from activation functions \footnote{We note that applying permutations to input weight vectors as in \cref{eq:sym} cannot be used as an augmentation technique when using invariant weight space networks.}. Notably these symmetries are much more difficult to incorporate directly into the weight space architectures compared to the permutation symmetries discussed in \autoref{sec:prelim}.
Here, we propose two augmentations that exploit this symmetry. 
When using the popular SIREN approach \citep{Sitzmann2020ImplicitNR}, the sinusoidal activation function induces two symmetries. First, since the function is odd ($\sin(x) = -\sin(x)$) we can multiply by $-1$ the weights and biases of a layer and the weights of the following layer. This gives the following new weights: $$\widetilde{W}_{i+1}=-W_{i+1},\widetilde{W}_{i}=-W_{i},\widetilde{b}_{i}=-b_{i}$$
This is justified due to the following equality $W_{i+1}\sin(W_ix{+}b) = -W_{i+1}\sin(-W_ix{-}b).$

The second symmetry results from the shift of the phase in an even or odd multiple of $\pi$, more formally for $k\in \mathbb{Z}$ we define:
$$\widetilde{W}_{i+1}=(-1)^kW_{i+1},\widetilde{W}_{i}=W_{i},\widetilde{b}_{i}=b_{i}+k\pi.$$
Again this gives functionally equivalent weights since:  $ W_{i+1}\sin(W_ix+b) = (-1)^kW_{i+1}\sin(W_ix+b+k\pi).$
We incorporate these symmetries through random data augmentations and refer to them as \textit{SIREN negation} and \textit{SIREN bias} respectively. We further discuss the ReLU activation symmetry in \cref{app:relu}.



\subsection{Weight space MixUp} \label{sec:MixUp}


In this section, we propose three weight space data augmentation schemes based on the MixUp method \cite{zhang2017mixup}. The main idea behind the standard MixUp method is to merge two samples into a new sample via a convex combination of both the inputs and the associated labels.

\textbf{Direct weight space MixUp.} Directly applying this idea to two weight vectors  $v_1=[W_l^{1}, b_l^{1}]$ and $v_2=[W_l^{2}, b_l^{2}]$, $l=1,\dots,M$ results in  what we call \emph{Direct weight space MixUp} as defined below. Formally, a sample from this MixUp method is a weight vector $v=[W_m,b_m]_{m=1}^M$ defined as follows:
\begin{equation} \label{eq:direct MixUp}
\begin{aligned}
W_l &= \lambda W_l^{1} + (1-\lambda) W_l^{2} \\
b_l &= \lambda b_l^{1} + (1-\lambda) b_l^{2} \\
y &= \lambda y^*_1 + (1-\lambda) y^*_2,
\end{aligned}
\end{equation}
where the weight parameter $\lambda$ is randomly drawn from some distribution (e.g., beta), and $y^*_i$ is the one-hot representation of $y_i$.

\textbf{The alignment problem.} A possible problem with direct weight space mixup is that it does not take weight space symmetries into account.  Specifically, the direct convex combination of two weight vectors might not be meaningful since the models are not properly aligned, that is, their weights and biases might not be positioned in a way that the corresponding entries in the weight matrices of these networks are similar. Formally, the weight alignment  \citep{entezari2022the, ainsworth2022git} refers to the combinatorial optimization problem for finding optimal permutations $p=(P_1,\dots, P_{M-1})$ that minimize 
$$\|v_1-p\cdot v_2\|_2,$$
and $p\cdot v_2$ applies the permutations to the weight vector $v_2$,  as defined in Equation \ref{eq:sym}.

\textbf{Alignment-based weight space MixUp.} A principled approach to deal with the aforementioned alignment problem would be to solve it, obtain an optimal alignment, and use it in the MixUp. As we discuss below this method can be nicely motivated by recent works on linear mode connectivity and model merging. Unfortunately, the alignment problem is NP-hard \cite{ainsworth2022git} and in practice, we need to use algorithms that approximate the optimal alignment. In this paper, we use the Weight Matching method from \cite{ainsworth2022git} \footnote{The algorithm iteratively solves simplified linear assignment problems based on the weight alignment functional.}, which is the fastest weight alignment algorithm that we were able to find. This gives rise to \emph{Alignment-based weight space MixUp}.  Formally, let $p^*(v_1,v_2)=(P^*_1,\dots,P^*_M)$ be the approximate alignment obtained from the weight matching algorithm, we define a new vector $v=[W_m,b_m]_{m=1}^M$ by the following equations:
\begin{equation} \label{eq:MixUp_with_perm}
\begin{aligned}
W_l &= \lambda W_{l}^{1} + (1-\lambda) P^{*}_{l} W_{l}^{2} P^{*T}_{l-1} \\
b_l &= \lambda b_{l}^{1} + (1-\lambda) P^{*}_{l} b_{l}^{2},
\end{aligned}
\end{equation}
where the labels are defined as in the previous equations.

Lastly, we present another variant of the Weight space MixUp:  

\textbf{Randomized weight space MixUp.} Another, simpler approach to partially address the alignment problem is to randomly apply permutations to the second weight vector. This gives rise to  \emph{Randomized weight space MixUp} which is defined as the mixup of a weight vector $v_1$ with a randomly permuted version of $v_2$. More formally, given a random sequence of permutations $p=(P_1,\dots, P_{M-1})$ sampled from some distribution, the new weight vector $v=[W_m,b_m]_{m=1}^M$ is defined as follows:
\begin{equation} \label{eq:MixUp_with_perm}
\begin{aligned}
W_l &= \lambda W_l^{1} + (1-\lambda) P_{l} W_l^{2} P_{l-1}^T \\
b_l &= \lambda b_l^{1} + (1-\lambda) P_{l} b_l^{2}, 
\end{aligned}
\end{equation}
and the labels are defined as in the previous equation.

\begin{table*}[t]
\scriptsize
\centering
\caption{\textit{INR classification}: test accuracy results for varying views.} 
\vspace{5pt}
\begin{tabular}{lccclccccc}
\toprule
Augmentation type & Model & \multicolumn{2}{c}{ModelNet40-INR} & & \multicolumn{2}{c}{FMNIST-INR} && \multicolumn{2}{c}{CIFAR10-INR}\\ \cmidrule(lr){3-4} \cmidrule(lr){6-7} \cmidrule(lr){9-10} 
 &  & 1 View & 10 View  && 1 View & 10 View &  & 1 View & 5 View\\

\midrule 
No augmentation  & DWS  &  $16.17 \pm 0.25$ & $30.25 \pm 0.95$  && $68.30 \pm 0.62$ & $76.01 \pm 1.20$  && $39.42 \pm 1.11$ & $40.64 \pm 0.22$\\
No augmentation  & GNN  &  $8.82 \pm 1.08$  & $34.51 \pm 1.24$  && $68.84 \pm 0.41$ & $79.58 \pm 3.01$ && $38.10 \pm 0.66$ & $46.52 \pm 1.53$\\
\midrule
Translate        &  DWS  &   $18.18 \pm 0.97$  & $31.17 \pm 0.02$  && $67.90 \pm 0.24$ & $77.61 \pm 0.36$ && $38.76 \pm 0.77$ & $40.95 \pm 0.39$\\
SIREN negation   &  DWS  &   $20.14 \pm 0.98$   & $32.31 \pm 0.70$  && $71.40 \pm 0.29$ & $77.71 \pm 1.38$ && $31.03 \pm 0.66$ & $29.15 \pm 0.67$\\
Masking          &  DWS  &   $11.43 \pm 2.44$   & $14.71 \pm 1.14$ && $68.48 \pm 0.14$ & $75.57 \pm 1.91$ && $36.48 \pm 0.89$ & $37.11 \pm 0.78$\\
Gaussian noise   &  DWS  &   $14.10 \pm 0.71$   & $25.31 \pm 1.78$ && $68.53 \pm 0.09$ & $77.60 \pm 0.13$ && $38.36 \pm 0.05$ & $38.54 \pm 0.57$\\
Translate        &  GNN  &   $8.17 \pm 0.81$   & $34.93 \pm 1.31$  && $70.17 \pm 1.26$ & $\textbf{83.83} \pm \textbf{0.25}$ && $38.06 \pm 0.81$ & $46.48 \pm 0.38$\\
SIREN negation   &  GNN  &   $11.41 \pm 3.22$   & $37.93 \pm 2.26$  && $72.74 \pm 4.29$ & $82.36 \pm 3.66$ && $35.80 \pm 1.02$ & $36.37 \pm 1.59$\\
Masking          &  GNN  &   $8.10 \pm 0.43$   & $18.04 \pm 1.24$ && $68.55 \pm 1.21$ & $79.72 \pm 1.35$ && $36.49 \pm 1.45$ & $42.69 \pm 0.29$\\
Gaussian noise   &  GNN  &   $9.06 \pm 0.27$   & $32.82 \pm 1.14$ && $77.55 \pm 0.33$ & $81.28 \pm 0.50$ && $45.66 \pm 0.35$ & $44.88 \pm 1.14$ \\
\midrule 
MixUp            &  DWS  &   $26.96 \pm 0.91$   & $31.92 \pm 0.37$  && $74.36 \pm 1.17$ & $78.58 \pm 0.20$ && $41.23 \pm 0.47$ & $40.94 \pm 0.26$\\
MixUp + random perm. &  DWS &  $26.62 \pm 0.18$   & $\textbf{33.55} \pm \textbf{1.40}$  && $73.89 \pm 0.89$ & $78.04 \pm 1.02$ && $38.34 \pm 0.54$ & $40.78 \pm 0.24$\\
Alignment + MixUp    &  DWS &  $\textbf{27.40} \pm \textbf{0.97}$   & $33.33  \pm 0.43$  && $\textbf{75.67} \pm \textbf{0.36}$ & $\textbf{79.41} \pm \textbf{0.56}$ && $\textbf{42.76} \pm \textbf{0.12}$ & $\textbf{43.36} \pm \textbf{0.40}$\\
MixUp            &  GNN  &   $20.45 \pm 3.82$   & $42.25 \pm 3.83$  && $\textbf{80.18} \pm \textbf{0.59}$ & $82.20 \pm 0.52$ && $\textbf{47.49} \pm \textbf{1.18}$ & $47.96 \pm 0.86$\\
MixUp + random perm. &  GNN &  $24.46 \pm 2.92$   & $41.67 \pm 4.55$  && $78.45 \pm 2.29$ & $82.24 \pm 0.68$ && $44.01 \pm 1.17$ & $45.36 \pm 0.67$\\
Alignment + MixUp    &  GNN &  $\textbf{26.88} \pm \textbf{1.75}$   & $\textbf{42.83} \pm \textbf{4.18}$  && $78.80 \pm 2.12$ & $82.94 \pm 0.31$ && $46.60 \pm 0.68$ & $\textbf{48.50} \pm \textbf{0.55}$\\
\bottomrule

\end{tabular}
\label{tab:exp_inr_cls}
\end{table*}

\textbf{Discussion.} 
Recent work \citep{entezari2022the, ainsworth2022git, pena2023re, navon2023equivariant} has shown that neural network weight vectors trained on the same loss function demonstrate a property called \emph{linear mode connectivity}. Specifically, interpolating between a weight vector $v_1$ and an optimally permuted version of another weight vector $p \cdot v_2$ results in intermediate weight vectors that maintain low loss values. In contrast, directly interpolating between $v_1$ and $v_2$ typically yields much higher loss values for the intermediate weights. This phenomenon suggests that linearly mixing aligned weight vectors preserves certain functional properties that lead to low loss values.  We note that when it comes to INRs, the loss mentioned above refers to the loss used when fitting the INRs. Therefore, when interpolating between INRs representing similar images, intermediate interpolation weights preserving the low loss should also represent a similar image. This might give additional motivation to use the alignment-based version of the weight space mixup.

\textbf{Relation to MixUp methods for other data types.} In the last few years, MixUp was successfully generalized to several data types such as point clouds \citep{chen2020pointMixUp,achituve2021self} and graphs \citep{han2022g,ling2023graph}.  There are some similarities in concept between these works and the alignment-based MixUp variant we propose, as point clouds and graphs both possess symmetries and alignment can be used to define an effective MixUp method for these data modalities as well.

\section{Experiments}\label{sec:exp}
We evaluate the various weight space augmentations we described on different datasets, learning setups, and weight space architectures. 
\revision{To support future research and the reproducibility of our results, we made our source code and datasets publicly available at: \url{https://github.com/AvivSham/deep-weight-space-augmentations}.} 

We report the average accuracy and standard deviations for $3$ random seeds.  Additional experimental results and details including the data generation processes are provided in the Appendix.

\textbf{Weight space architectures}. 
In the following sections, we examine the effect of weight space augmentations on the performance of two leading weight space architectures.

\textit{DWS}~\cite{navon23dws} is a weight space architecture that is equivariant to permutation symmetries imposed by neural weights. This architecture is composed of a series of permutation equivariant layers, interleaved with pointwise non-linear activations.

\textit{GNN}~\cite{zhang2023neural} transforms the input neural network into a graph that preserves the symmetries in the weight space. It comes with the advantage that we can directly leverage existing strong architectures for graphs like Principal Neighbourhood Aggregation \citep{corso2020pna} and transformers using Relational Attention \citep{diao2023relational}.

\revision{In all experiments, both DWS and GNN are equipped with weight decay and dropout for regularization. We use the values mentioned in the original papers.}

\textbf{Augmentations}. 
We compare the augmentations presented in Sections~\ref{sec:taxonomy} and \ref{sec:MixUp} to training without augmentations. These include: (1)~\textit{Translate} -- translating the image space data through the weight space. (2)~\textit{SIREN negation} -- exploiting the natural symmetry of $\sin(x) = -\sin(-x)$. (3)~\textit{MixUp} -- applying MixUp directly on a pair of weights and biases vectors. (4)~\textit{MixUp + random permutation} -- applying random permutation before performing MixUp. (5)~\textit{Alignment + MixUp} -- aligning the weights and biases vectors prior to MixUp. (6)~\textit{Masking} -- randomly zero out elements in the neural view. (7)~\textit{Gaussian Noise} -- adding gaussian distributed random noise to the neural view.
We present extended results with additional augmentations in Appendix~\ref{app:inr_classification}. 

\subsection{INR classification} \label{sec:inr_classification}
We evaluate the aforementioned weight space augmentations on implicit neural representation (INR) classification tasks. Specifically, we create INR datasets for ModelNet40 (3D point clouds), FMNIST (2D greyscale images), and CIFAR10 (2D RGB images). For each instance in the original dataset, we generate multiple views by sampling different weight initializations. We generate 10, 10, and 5 views for FMNIST, ModelNet40, and CIFAR10 respectively.
Table \ref{tab:exp_inr_cls} demonstrates the effectiveness of on-the-fly weight space data augmentation schemes. Notably, MixUp augmentations with a single view are comparable to training with 5--10x more data across the INR datasets: ModelNet40, FMNIST, and CIFAR10. Furthermore, data augmentation is still effective when training with 10 views.  \cref{fig:MixUp_effect} depicts a training curve with and without the use of our alignment-based MixUp technique. It is evident that MixUp mitigates overfitting and improves performance. 
Overall, the effectiveness of input space and data-agnostic augmentations varies between models and datasets and is not as significant as the improvements from using weight space MixUp.

\textbf{Comparison of Mixup variants.} All variants of the weight space MixUp provide consistent improvements, with the alignment-based version frequently outperforming other MixUp variants by 1-3\%. For DWS, the alignment-based MixUp is the best among all augmentation types in five out of six setups in Table \ref{tab:exp_inr_cls} and comparable in the last setup. For GNNs, the alignment-based MixUp is the best in three out of six cases, where interestingly the direct MixUp variant provides strong competition. The randomized mixup version is worse than the other approaches in many cases. We note that this somewhat underwhelming advantage of the alignment-based MixUp is surprising. One possible reason is that our weight alignment algorithm might produce suboptimal alignments. This can be due to the algorithm's inability to produce good permutation alignments, and also since it only considers permutation symmetries and overlooks the additional symmetries such as activation symmetries.
We anticipate that employing a higher-quality alignment method that respects both permutations and negations symmetries would address this issue and consequently lead to improved performance. One possible way to obtain such an alignment method is to learn it directly from data \cite{navon2023equivariant}.
We conducted complementary experiments that compare MixUp variants with feature MixUp, in addition to an ablation study assessing the contribution of the key elements of MixUp, namely label smoothing and input averaging. The detailed results of these experiments can be found in \cref{app:feature_mixup}, \ref{app:mixup_ablation}.

\begin{table*}[t]
    \centering
\caption{Semi supervised classification: test accuracy results across different levels of labeled data.}
\scriptsize
\label{tab:ssl}
    \begin{tabular}{lcccccccc}
    \toprule
   & \multicolumn{4}{c}{FMNIST} & \multicolumn{4}{c}{CIFAR10}\\
   \cmidrule(lr){2-5} \cmidrule(lr){6-9}
           &  5\%&  10\%&  15\%& 20\% & 5\%& 10\%& 15\%&20\% \\
           \midrule
 DWS& $48.55 \pm 0.59$& $54.01 \pm 0.61$& $59.27 \pm 0.56$& $57.60 \pm 0.74$& $22.95 \pm 0.82$& $25.48 \pm 0.96$& $26.36 \pm 0.55$ & $28.30 \pm 0.58$\\
         DWS + SSL&  $\textbf{59.49} \pm \textbf{0.69}$ &  $\textbf{61.68} \pm \textbf{0.15}$ &  $\textbf{63.34} \pm \textbf{0.44}$ &  $\textbf{64.67} \pm \textbf{0.54}$ & $\textbf{27.63} \pm \textbf{1.13}$ & $\textbf{30.08} \pm \textbf{1.29}$ & $\textbf{31.75} \pm \textbf{1.10}$ & $\textbf{32.75} \pm \textbf{0.36}$ \\
         \midrule
         GNN  & $47.43 \pm 2.44$ & $57.58 \pm 1.51$ & $59.76 \pm 1.14$ & $61.56 \pm 0.84$ & $21.07 \pm 1.29$ & $23.51 \pm 0.58$ & $25.67 \pm 0.22$ & $27.52 \pm 0.85$\\
           GNN + SSL & $\textbf{55.58} \pm \textbf{1.81}$ & $\textbf{60.18} \pm \textbf{0.67}$ & $\textbf{62.09} \pm \textbf{0.59}$ & $\textbf{62.84} \pm \textbf{0.82}$ & $\textbf{26.93} \pm \textbf{1.08}$ & $\textbf{30.05} \pm \textbf{0.49}$ & $\textbf{30.96} \pm \textbf{0.66}$ & $\textbf{32.69} \pm \textbf{0.85}$\\
           \bottomrule
    \end{tabular}

\end{table*}

\subsection{Representation learning in weight spaces} \label{sec:ssl}
%
While we have shown the utility of weight space augmentations for supervised learning, they may also prove beneficial for other paradigms like unsupervised representation learning.
To test this, we conduct an experiment where we use contrastive learning to pre-train a weight space model and evaluate the learned representations on downstream INR classification tasks. Specifically, in the training stage, we use the SimCLR contrastive learning framework \cite{chen2020simple} where positive examples are generated by applying a sequence of augmentations including translation, Gaussian noise, quantile masking, SIREN negation, and alignment + MixUp. After training, we evaluate the model by adding an MLP classification head to the pre-trained model and fine-tuning the whole model end-to-end on a small labeled subset of the training data. Experiments are performed on FMNIST and CIFAR10 INR datasets. We compare against a natural baseline trained directly and only on the labeled data. The fraction of labeled examples used for fine-tuning is varied from 5\% to 20\% of the full training set.

\begin{figure}[t]
    \centering
\includegraphics[width=0.9\linewidth, trim={0 4mm 0 0}, clip]{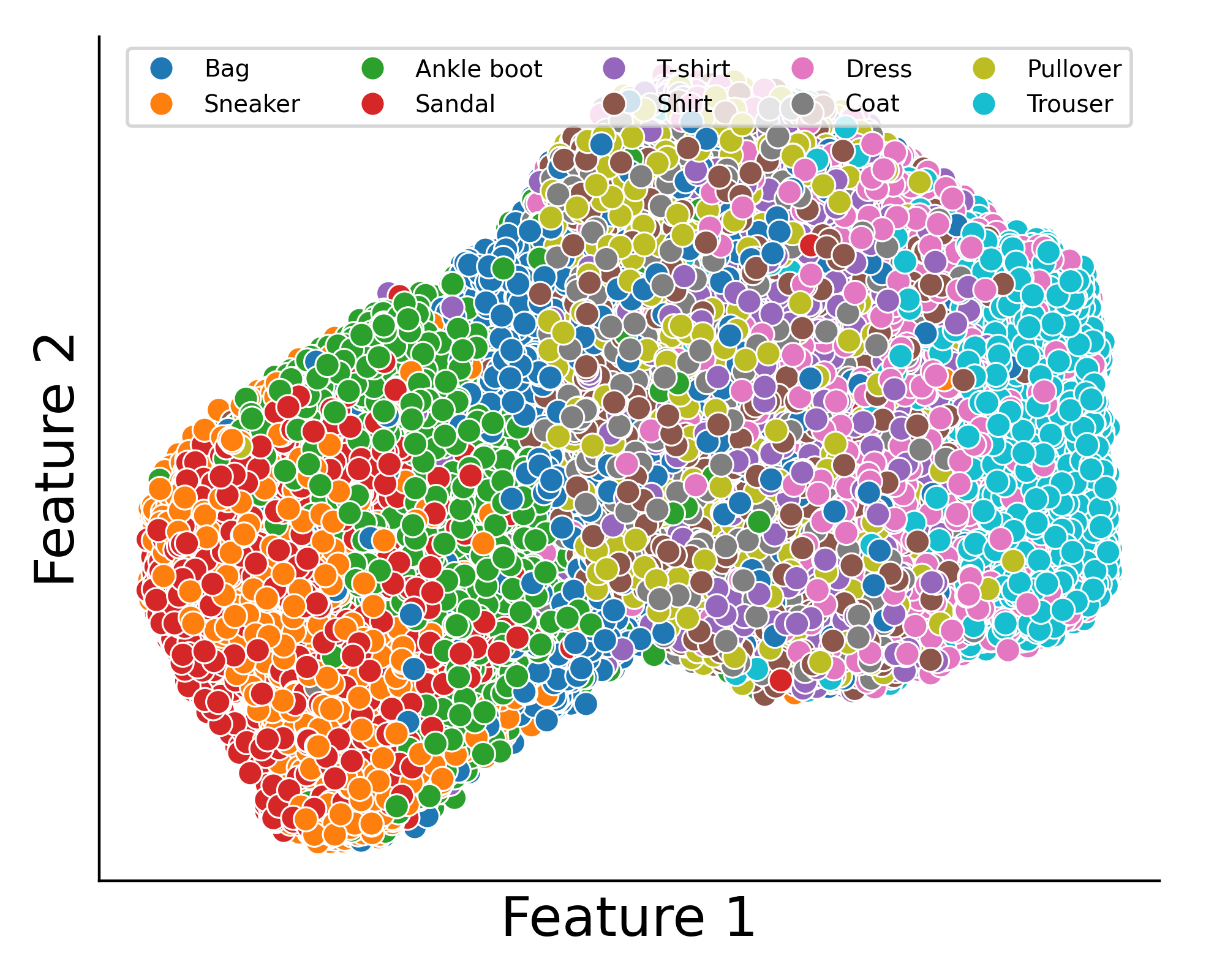}
    \caption{\textit{Representations in weight space}: Visualization of the 2D feature space attained through SimCLR contrastive learning.}
    \label{fig:ssl_rep}
\end{figure}

The results are presented in Table~\ref{tab:ssl}. We show that using weight space augmentations for pre-training weight space architectures with contrastive learning improves the model's performance. The pre-trained models, DWS+SSL and GNN+SSL consistently outperform their fine-tune-only versions across all datasets as well as with respect to the variations in the number of labeled training examples in the fine-tuning step. Specifically, DWS+SSL and GNN+SSL significantly improve performance by up to ${\sim}11\%$ and ${\sim}8\%$ of accuracy respectively in comparison to their fine-tune-only counterparts. This finding indicates that a meaningful representation is learned in the pre-training stage. In addition, we visualize the representations learned by the feature extractor by performing dimensionality reduction using UMAP~\cite{mcinnes2018umap}. The results are presented in Figure~\ref{fig:ssl_rep}. The figure shows that the feature extractor learns meaningful representations in two ways: (1) there is a clear distinction between different classes. (2) clothing items of the same category (e.g. shoes) exhibit proximity in the representation space. Additional qualitative comparison to augmentations proposed by \citet{navon23dws} can be found in Appendix~\ref{app:representation}.

\section{Previous work}\label{sec:prev}

\textbf{Learning in deep weight spaces.}
Learning in deep weight spaces is a relatively new learning setup, where we use a special type of neural network to process the parameters (weights and biases) of \emph{other neural networks}. Pioneering works \citep{eilertsen2020classifying,unterthiner2020predicting,andreis2023set,herrmann2023learning, Wang2022DeepLO, Xu2022SignalPF} in this direction suggested using standard architectures like fully connected networks or transformers for this setup. One limitation is that they disregard the permutation symmetry structure of weight spaces (see \cref{sec:prelim}). As symmetries provide meaningful inductive biases, several recent works suggested \emph{weight space} architectures that are equivariant to these permutation symmetries  \citep{navon23dws,zhou2023permutation,zhou2023neural,zhang2023neural,lim2023graph}.  These works have shown an improvement over previous approaches in INR classification and editing tasks, but on most benchmarks, their performance still significantly lags behind standard architectures applied to the original data signals.

\textbf{Implicit Neural Representations.} 
In this paper we focus on classifying Implicit Neural Representations (INRs) \cite{Sitzmann2020ImplicitNR,mescheder2019occupancy,park2019deepsdf} as the weight space task. INRs are an emerging type of data representation for objects such as 3D shapes, 3D scenes, and images. The core concept is to use a neural network to represent an object as a continuous function. For instance, an INR can take the form of a multilayer perceptron (MLP) that takes 2D spatial coordinates $(x,y)$ as input and outputs a grayscale value corresponding to this position in the image, see \cref{fig:overview}. Similarly, a neural network can be fitted to take 3D coordinates as input and output the signed distance to a surface to implicitly represent a 3D shape -- the zero-level set of this INR defines the original surface. More complex 3D scenes can also be represented in this implicit way using NeRFs \cite{mildenhall2021nerf}. Recent research has shown that INRs are capable of representing complex scenes and other objects faithfully, and they are considered an extremely important research direction, however, it remains unclear what type of neural network architecture is best suited to handle them. While we focus on INRs in this paper, the augmentations we propose can be applied to other types of functions represented by neural networks.

\section{Conclusion}
In this work, we explore overfitting in weight space architectures and introduce weight space augmentations to address this problem and improve model performance. Through experiments, we first examine key questions related to generalization behavior in weight space models and show that multiple neural views are an effective approach for improving generalization. We then review and design augmentation techniques leveraging inherent weight space structure. In particular, we propose three MixUp schemes for weight spaces. Our results demonstrate that training with these augmentations yields comparable performance to substantially expanding the training set size. Moreover, we find that weight space augmentations can be used effectively when conducting self-supervised learning in weight space.

\textbf{Limitations.}
 Despite the improved performance discussed above, a notable gap persists when compared to directly working in image or point cloud representations. Future research is needed in order to further close this gap. Additionally, the evaluation of data augmentation schemes in this paper centers on INRs. It is worth noting that the use of weight space augmentations may be extended to other scenarios, including generalization prediction and learning to optimize \citep{zhang2023neural}. Further research is required to evaluate our data augmentation techniques and to develop novel data augmentation techniques for this setup.


\section*{Impact Statement}
This paper presents work whose goal is to advance the field of Machine Learning. There are many potential societal consequences of our work, none which we feel must be specifically highlighted here.

\section*{Acknowledgements.} HM is the Robert J. Shillman Fellow, and is supported by the Israel Science Foundation through a personal grant (ISF 264/23) and an equipment grant (ISF 532/23). This study was funded by a grant to GC from the Israel Science Foundation (ISF 737/2018), by an equipment grant to GC and Bar-Ilan University from the Israel Science Foundation (ISF 2332/18), and by the Israeli Ministry of Science, Israel-Singapore binational grant 207606. AN and AS are supported by a grant from the Israeli higher-council of education, through the Bar-Ilan data science institute (BIU DSI).


\bibliography{ref}
\bibliographystyle{icml2024}

\newpage
\appendix
\onecolumn

\section{Datasets}
\label{app:datasets}
The increasing usage of INRs in many machine-learning domains, specifically in images and 3D objects, raises the need for INR benchmarks. Implicit representations, such as neural radiance fields and neural implicit surfaces, offer a more flexible and expressive way to model complex 3D scenes and objects. However, as these techniques gain traction, it becomes crucial to establish standardized benchmarks to assess and compare the performance of architectures designed for weight space data. To address this issue, we present new INR classification benchmarks based on ModelNet40~\cite{wu20153d}, Fashion-MNIST~\cite{Xiao2017FashionMNISTAN}, and CIFAR10~\cite{krizhevsky2009learning} datasets. We use the SIREN~\citep{Sitzmann2020ImplicitNR} architecture, i.e. MLP with sine activation, and fit each objects in the original dataset. To negate the possibility of canonical representation that may lead to globally aligned data representation, we randomly initialize the weights for every generated INR. 
In the case of ModelNet40, INRs are generated through training an MLP to accurately predict the signed distance function values of a 3D object given a set of 3D point clouds. For Fashion-MNIST and CIFAR10 an MLP is trained to map from the 2D xy-grid to the corresponding gray or RGB intensity level value in the original image. 
We fit $10$ unique INRs, namely views, per sample in the original dataset for ModelNet40 and Fashion-MNIST respectively. For CIFAR10 dataset we fit $5$ unique INRs per image.

\section{Datasets generation}

\emph{Fashion-MNIST INRs.} We fit an INR to each image in the original dataset. We split the INRs dataset into train, validation, and test sets of sizes 55K, 5K, and 10K respectively. Each INR is a $5$-layer MLP network with a $32$ hidden dimension, i.e., $3 \xrightarrow{} 32 \xrightarrow{} 32 \xrightarrow{} 32 \xrightarrow{} 32 \xrightarrow{} 3$. We train the INRs using the Adam optimizer for $10K$ steps with a learning rate of $5e-4$. When the PSNR of the reconstructed image from the learned INR is greater than $40$, we use early stopping to reduce the generation time.

\emph{CIFAR10 INRs.} We fit an INR to each image in the original dataset. We use the train test splits presented by ~\cite{krizhevsky2009learning}. Each INR is a $3$-layer MLP network with a $32$ hidden dimension, i.e., $3 \xrightarrow{} 32 \xrightarrow{} 32 \xrightarrow{} 1$. We train the INRs using the Adam optimizer for $1K$ steps with a learning rate of $5e-4$. When the PSNR of the reconstructed image from the learned INR is greater than $40$, we use early stopping to reduce the generation time.

\emph{ModelNet40.}
We use the original split presented in~\cite{wu20153d} and fit an INR for each data sample. We start by converting the mesh object to a signed distance function (SDF) by sampling $250K$ points near the surface. Next, we fit a 5-layer INR with a hidden dim of $32$, i.e., $3 \xrightarrow{} 32 \xrightarrow{} 32 \xrightarrow{} 32 \xrightarrow{} 32 \xrightarrow{} 1$ by solving a regression problem. Given a 3 dimensional input, the INR network predicts its SDF. For the optimization, we use AdamW optimizer with $1e-4$ learning rate and perform $1000$ update steps.

\section{Weight space augmentations} \label{app:augs_info}

\begin{wrapfigure}[15]{r}{0.5\textwidth}
\vspace{-20pt}
\begin{center}
    \includegraphics[width=0.49\textwidth]{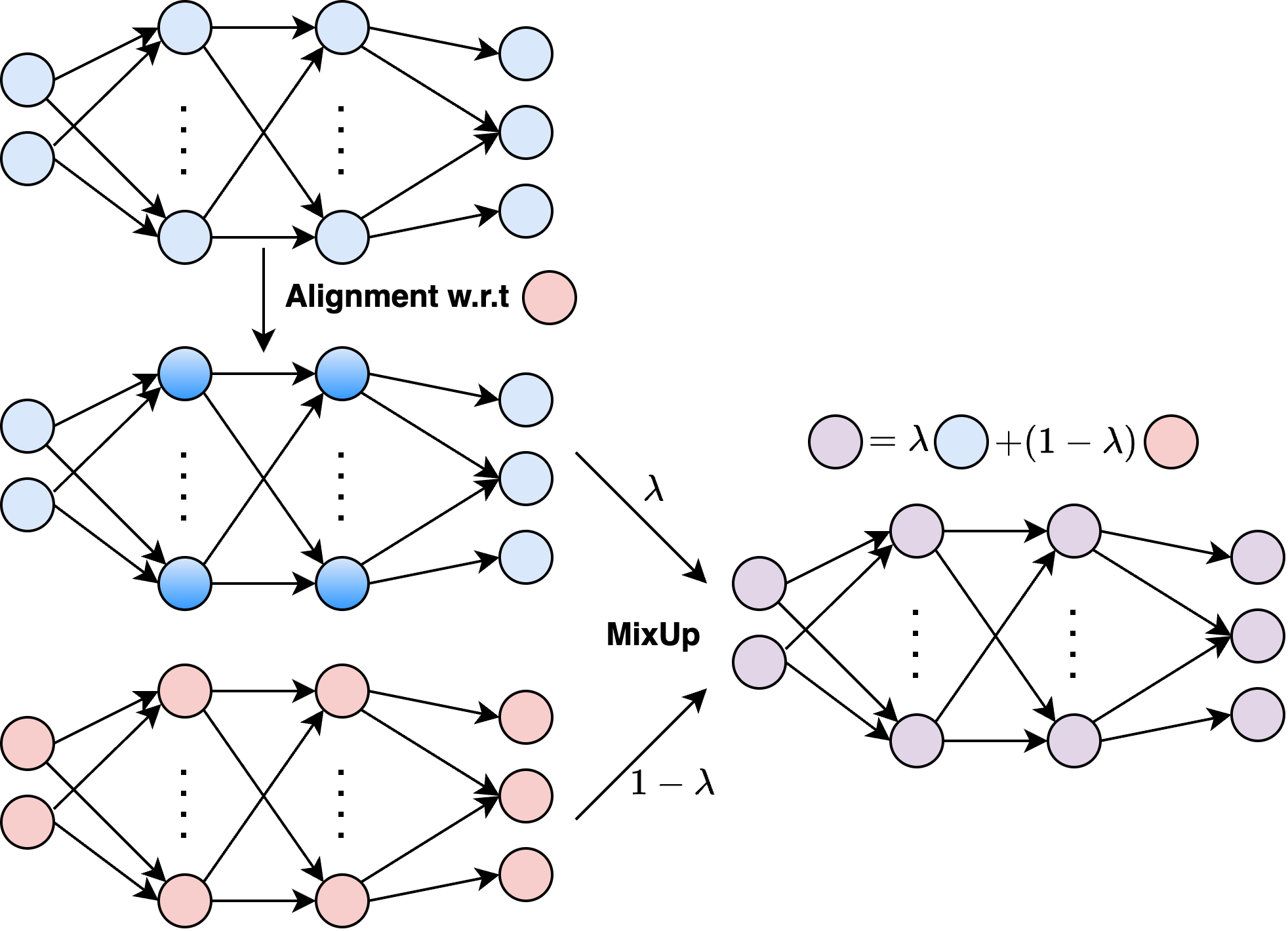}
  \end{center}
\caption{\textit{Illustration of alignment + mixup augmentation}.}
\label{fig:mixup_alignment}
\end{wrapfigure}

We provide a detailed description and hyperparameter choice we use in the experimental sections.

\textit{Translate} - we translate the weights of INR by drawing translate parameter $t$ from uniform distribution, i.e. $t \sim \mathcal{U}(-u, u)$. In our experiments, we used $u=0.25$ 

\textit{Rotation} - we apply 2D rotation matrix $R$ on the first weight matrix of the INR, i.e. $W_1 = RW_1$. We draw the rotation angle $\varphi$ from a uniform distribution $\varphi \sim \mathcal{U}(-u, u)$. Specifically, we set $u=30$ in degrees unit. 

\textit{Scale} - We apply random scaling to the weight metrics using a random variable $c$, where $c \sim \mathcal{U}(0.8, 1)$.  

\textit{SIREN bias} - We exploit the natural symmetry of sine function $\sin(x+k\pi) = (-1)^k\sin(x)$ and apply it to the input set of weights and biases. More formally,
$W_{i+1}\sin(W_ix+b) = (-1)^kW_{i+1}\sin(W_ix+b+k\pi)$, for $k\in\mathbb{Z}$.

\textit{SIREN negation} - We exploit the natural symmetry of sine function $\sin(x) = -\sin(-x)$. Since the function is odd we can negate the weight and biases of layer $i$ and the weight of the following layer $i{+}1$ as $W_{i+1}\sin(W_ix+b) = -W_{i+1}\sin(-W_ix-b)$.

\textit{Masking} - randomly zero out elements from the neural view according to a pre-defined masking rate. In our experiments, we set the masking rate to $0.1$, i.e., $10\%$ of the elements mapped to zero.

\textit{Quantile Masking} - acts similarly to \textit{Masking} augmentation, but instead of randomly zeroing the elements, here we zero elements with magnitudes below a threshold. In our experiments, we randomly draw the threshold from $\mathcal{U}(0, 0.1)$.

\textit{Gaussian Noise} - adding additive noise drawn from Gaussian distribution to the neural view. More formally, the noise is drawn from $\mathcal{N}(0, s\sigma_{W_i})$, where $s$ denotes scaling factor and $\sigma_{W_i}$ the standard deviation of $W_i$. In our experiments, we set $s=0.32$.

\textit{Combination} - here, we sequentially apply a combination of weight space augmentations. This combination includes translate, gaussian noise, SIREN negation, and alignment + MixUp. Note that we did not experiment with all possible combinations as it grows exponentially and is highly computationally demanding.

\subsection{ReLU scaling augmentation} \label{app:relu}
We present additional weight space augmentation tailored for ReLU activation symmetry, as well as other homogeneous activations (i.e. $f(\alpha x) = \alpha f(x)$). We can arbitrarily scale the weights of any layer and define new weights by 
$$\widetilde{W}_{i} = CW_{i}, \widetilde{b}_{i} = Cb_i, \widetilde{W}_{i+1} = W_{i+1}C^{-1}, $$
where $C$ is a diagonal matrix with positive entries. 
These new weights are functionally equivalent to the original weights since ReLU is equivariant to multiplication with positive diagonal matrices:  $ W_{i+1}C^{-1}\text{ReLU}(CW_ix{+}Cb_i){+}b_{i{+}1} = W_{i+1}\text{ReLU}(W_ix{+}b_i){+}b_{i+1}$.

\begin{table}[t]
\scriptsize
\centering
\caption{\textit{INR classification:} ModelNet40, FMNIST, and CIFAR10 test accuracy results with varying number of views.} 
\vspace{5pt}
\begin{tabular}{lccccccccc}
\toprule
Augmentation type & Model & \multicolumn{2}{c}{ModelNet40} &  & \multicolumn{2}{c}{FMNIST} &  & \multicolumn{2}{c}{CIFAR10}\\ \cmidrule(lr){3-4} \cmidrule(lr){6-7} \cmidrule(lr){9-10} 
 &  & 1 View & 10 View &  & 1 View & 10 View &  & 1 View & 10 View\\

\midrule 
No augmentation  & DWS  &  $16.17 \pm 0.25$ & $30.25 \pm 0.95$ && $68.30 \pm 0.62$ & $76.01 \pm 1.20$ && $40.64 \pm 0.22$ & $40.64 \pm 0.22$\\
No augmentation  & GNN  &  $8.82 \pm 1.08$  & $34.51 \pm 1.24$ && $68.84 \pm 0.41$ & $79.58 \pm 3.01$ && $38.11 \pm 0.66$ & $46.52 \pm 0.15$\\
\midrule
Translate        &  DWS  &   $18.18 \pm 0.97$  & $31.17 \pm 0.02$ && $67.90 \pm 0.24$ & $77.61 \pm 0.36$ && $38.76 \pm 0.77$ & $40.95 \pm 0.39$\\
Rotation         &  DWS  &   ---   & --- && $68.55 \pm 0.28$ & $77.04 \pm 0.47$ && $39.21 \pm 0.61$ & $40.46 \pm 0.08$ \\
Scale            &  DWS  &   $16.41 \pm 0.57$   & $30.54 \pm 0.72$ && $67.99 \pm 0.14$ & $75.77 \pm 1.09$ && $39.34 \pm 0.91$ & $40.93 \pm 0.49$ \\
Gaussian noise   &  DWS  &   $14.10 \pm 0.71$   & $25.31 \pm 1.78$ && $68.53 \pm 0.09$ & $77.60 \pm 0.13$ && $38.36 \pm 0.05$ & $38.54 \pm 0.57$\\
SIREN bias       &  DWS  &   $4.69 \pm 0.10$   & $4.90 \pm 0.01$ && $58.20 \pm 0.01$ & $62.21 \pm 0.55$ && $24.34 \pm 0.56$ & $22.83 \pm 0.45$\\
SIREN negation   &  DWS  &   $20.14 \pm 0.98$   & $32.31 \pm 0.70$ && $71.40 \pm 0.29$ & $77.71 \pm 1.38$ && $31.03 \pm 0.66$ & $29.15 \pm 0.67$\\
Masking          &  DWS  &   $11.43 \pm 2.44$   & $14.71 \pm 1.14$ && $68.48 \pm 0.14$ & $75.57 \pm 1.91$ && $36.48 \pm 0.89$ & $37.11 \pm 0.78$\\
Quantile masking &  DWS  &   $15.13 \pm 2.45$   & $29.88 \pm 0.62$ && $68.72 \pm 0.27$ & $76.22 \pm 0.72$ && $39.46 \pm 0.35$ & $40.61 \pm 0.21$\\
Translate        &  GNN  &   $8.17 \pm 0.81$   & $34.93 \pm 1.31$ && $70.17 \pm 1.26$ & $\mathbf{83.83} \pm \mathbf{0.25}$ && $38.06 \pm 0.81$ & $46.41 \pm 0.17$\\
Rotation         &  GNN  &   ---   & --- && $69.35 \pm 2.18$ & $83.72 \pm 1.14$ && $36.65 \pm 0.74$ & $46.61 \pm 0.44$\\
Scale            &  GNN  &   $8.58 \pm 0.65$   & $34.70 \pm 5.19$ && $68.96 \pm 1.46$ & $83.67 \pm 0.19$ && $38.23 \pm 0.65$ & $46.82 \pm 0.74$\\
Gaussian noise   &  GNN  &   $9.06 \pm 0.27$   & $32.82 \pm 1.14$ && $77.55 \pm 0.33$ & $81.28 \pm 0.50$ && $45.66 \pm 0.35$ & $44.88 \pm 1.14$ \\
SIREN bias       &  GNN  &   $11.63 \pm 2.48$   & $34.32 \pm 1.57$ && $68.09 \pm 0.49$ & $77.20 \pm 1.03$ && $35.64 \pm 0.16$ & $35.84 \pm 0.94$\\
SIREN negation   &  GNN  &   $11.41 \pm 3.22$   & $37.93 \pm 2.26$ && $72.74 \pm 4.29$ & $82.36 \pm 3.66$ && $35.81 \pm 1.01$ & $36.37 \pm 0.16$\\
Masking          &  GNN  &   $8.10 \pm 0.43$   & $18.04 \pm 1.24$ && $68.55 \pm 1.21$ & $79.72 \pm 1.35$ && $36.49 \pm 1.45$ & $42.69 \pm 0.29$\\
Quantile masking &  GNN  &   $8.12 \pm 0.85$   & $34.36 \pm 1.14$ && $69.96 \pm 2.08$ & $83.78 \pm 0.76$ && $46.73 \pm 0.11$ & $38.47 \pm 0.12$ \\
\midrule 
MixUp            &  DWS  &   $26.96 \pm 0.91$   & $31.92 \pm 0.37$ && $74.36 \pm 1.17$ & $78.58 \pm 0.20$ && $41.23 \pm 0.47$ & $40.94 \pm 0.26$\\
MixUp + random perm. &  DWS &  $26.62 \pm 0.18$   & $\mathbf{33.55} \pm \mathbf{1.40}$ && $73.89 \pm 0.89$ & $78.04 \pm 1.02$ && $38.34 \pm 0.54$ & $40.78 \pm 0.24$\\
Alignment + MixUp    &  DWS &  $\mathbf{27.40} \pm \mathbf{0.97}$   & $33.33  \pm 0.43$ && $\mathbf{75.67} \pm \mathbf{0.36}$ & $\mathbf{79.41} \pm \mathbf{0.56}$ && $\mathbf{42.76} \pm \mathbf{0.12}$ & $\mathbf{43.36} \pm \mathbf{0.40}$\\
MixUp            &  GNN  &   $20.45 \pm 3.82$   & $42.25 \pm 3.83$ && $\mathbf{80.18} \pm \mathbf{0.59}$ & $82.20 \pm 0.52$ && $47.14 \pm 0.95$ & $47.56 \pm 0.58$\\
MixUp + random perm. &  GNN &  $24.46 \pm 2.92$   & $41.67 \pm 4.55$ && $78.45 \pm 2.29$ & $82.24 \pm 0.68$ && $43.97 \pm 0.86$ & $45.36 \pm 0.67$\\
Alignment + MixUp    &  GNN &  $\mathbf{26.88} \pm \mathbf{1.75}$   & $\mathbf{42.83} \pm \mathbf{4.18}$ && $78.80 \pm 2.12$ & $82.94 \pm 0.31$ && $\mathbf{47.66} \pm \mathbf{0.86}$ & $\mathbf{49.50} \pm \mathbf{0.55}$ \\
\midrule
Combination & DWS & $29.05 \pm 1.19$ & $35.60 \pm 1.20$ && $74.43 \pm 0.11$ & $78.57 \pm 0.47$ && $29.63 \pm 0.87$ & $32.01 \pm 0.65$\\
Combination & GNN & $13.65 \pm 0.68$ & $46.07 \pm 0.59$ && $75.36 \pm 1.57$ & $80.13 \pm 1.82$ && $37.46 \pm 0.57$ & $39.14 \pm 0.54$\\
\bottomrule

\end{tabular}
\label{tab:inrs_appendix}
\end{table}

\section{Experimental settings}
\label{app:exp_settings}
\paragraph{DWS.}
In all experiments, we use DWS~\cite{navon2023equivariant} network with $4$ hidden layers and hidden dimension of $128$. We optimized the network using a $5e-3$ learning rate with AdamW~\cite{Loshchilov2017FixingWD} optimizer.

\paragraph{GNN.}
For the GNN, we use the version of Relation Transformer presented in \cite{zhang2023neural} with 4 hidden layers, node dimension of $64$, and edge dimension of $32$. We optimized the network using a $1e-3$ learning rate with AdamW~\cite{Loshchilov2017FixingWD} optimizer and a $1000$ steps warmup schedule.

\paragraph{INR classification.}
We optimized the weight space architecture for $250$ epochs for the ModelNet40, and $300$ epochs for the FMNIST and CIFAR10 INRs datasets.  Additionally, we utilize the validation set for early stopping, i.e. selecting the best model w.r.t validation accuracy.

\paragraph{Representation learning.}
In this learning setup, we employ FMNIST and CIFAR10 INRs datasets to perform semi-supervised learning for INR classification.
We start by optimizing weight space architecture using the SimCLR framework as the pre-training step. Next, we fine-tune the model using a limited amount of labeled data. We use $100/400$ epochs in the pertaining step for FMNIST and CIFAR10 respectively, followed by $400$ epochs for fine-tuning using labeled data.

\begin{figure*}[t]
  \begin{subfigure}[Our augmentations.]{
    \includegraphics[width=0.45\linewidth]{figs/ssl_rep.png}
    }
    \label{fig:cifar10_mlp}
  \end{subfigure}  
  \hspace*{\fill}
  \begin{subfigure}[Augmentations from \cite{navon23dws}.]{
  \includegraphics[width=0.45\linewidth]{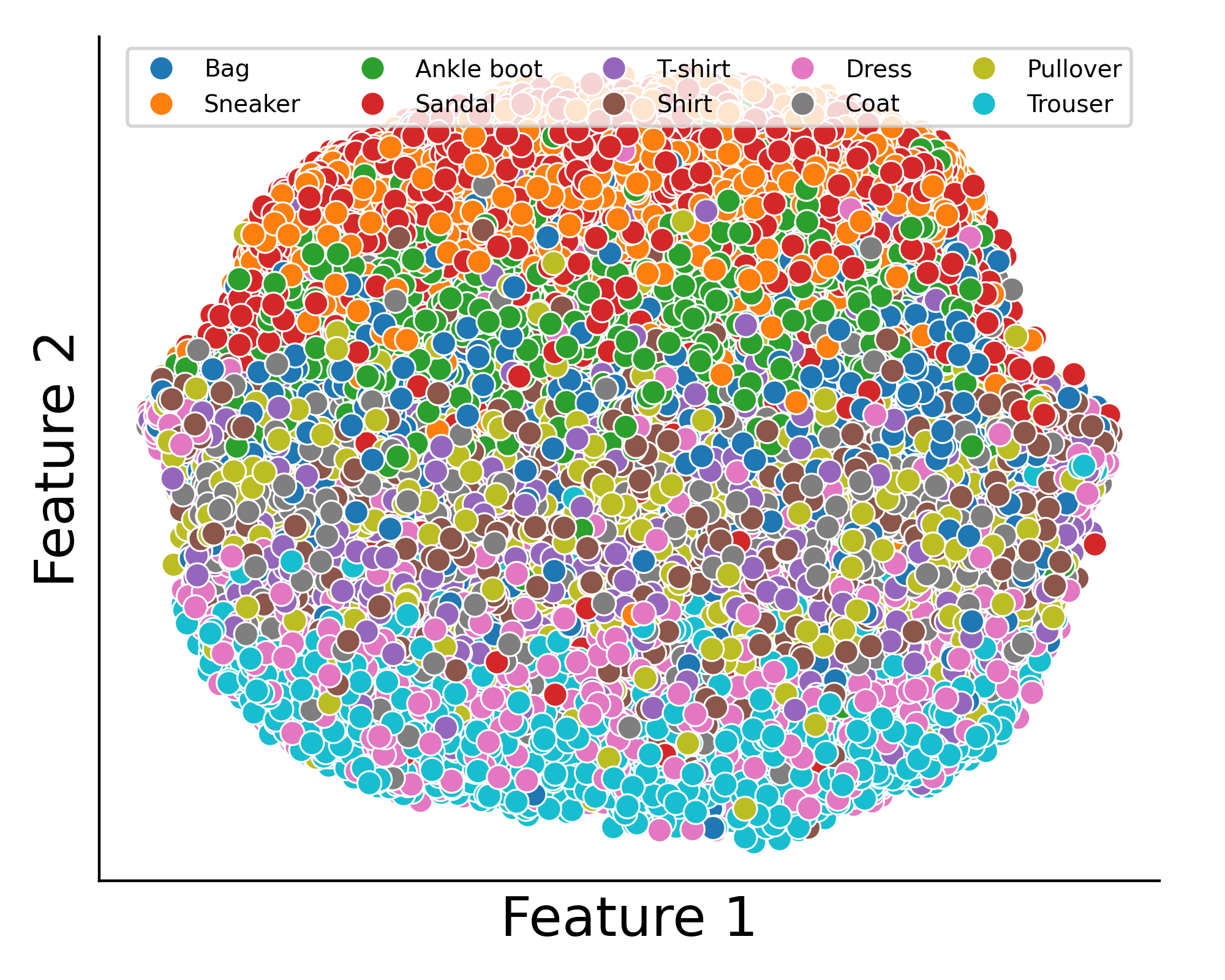}
    }
    \label{fig:ssl_comparison}
  \end{subfigure}
  \caption{\textit{Representations in weight space:} Visualization of the 2D feature space attained through SimCLR contrastive learning, showcasing the impact of our augmentations on the learned representations.}
 \label{fig:ssl_qualit}
\end{figure*}

\section{Additional experimental results}
\subsection{INR classification.}\label{app:inr_classification}
We present additional results complementing those discussed in Section~\ref{sec:inr_classification}, which encompass augmentations detailed in Appendix~\ref{app:augs_info}. We run every augmentation with $3$ random seeds and report the average accuracy and standard deviation. 

\subsection{Comparison to non-invariant architectures.} \revision{In this experiment, we compare DWS and GNN models, incorporating Alignment + MixUp augmentation, to architectures that are not invariant to permutation symmetries. Specifically, we use the transformer architecture proposed by~\cite{schurholt2021self} and re-run it twice: once with random permutation augmentation and once without. We evaluate the models on the FMNIST INR classification task for both 1 and 10 views.
The results are shown in Table~\ref{tab:non-invariant-appendix}. As demonstrated in~\cite{navon2023equivariant}, architectures that do not respect the permutation symmetries perform significantly worse than permutation equivariant models.}

\begin{table}[]
\centering
\caption{\textit{Comparison to non-invariant models:} FMNIST test accuracy results with varying number of views.} 
\vspace{5pt}
\small
\begin{tabular}{lccc}
\toprule
\multicolumn{1}{c}{Model} & Augmentation &  &  \\ \midrule
 &  & 1 View & 10 View \\ \midrule
Transformer & No augmentation & $27.60 \pm 0.25$ & $29.91 \pm 0.31$ \\
Transformer & Random permutation & $28.93 \pm 0.43$ & $30.14 \pm 0.06$ \\
\midrule
DWS & No augmentation & $68.30 \pm 0.62$ & $76.01 \pm 1.20$ \\
DWS & Alignment + MixUp & $\mathbf{75.67} \pm \mathbf{0.36}$ & $\mathbf{79.41} \pm \mathbf{0.56}$ \\
\midrule
GNN & No augmentation & $68.84 \pm 0.41$ & $79.58 \pm 3.01$ \\
GNN & Alignment + MixUp & $\mathbf{78.80} \pm \mathbf{2.12}$ & $\mathbf{82.94} \pm \mathbf{0.31}$ \\
\bottomrule
\end{tabular}
\label{tab:non-invariant-appendix}
\end{table}

\subsection{Representation learning in weight spaces.}\label{app:representation}
In this section, we provide additional results for the experiment presented in Section~\ref{sec:ssl}. We repeat the experiment again but with variations in the augmentations applied during SimCLR optimization. Specifically, we apply the augmentations presented in ~\cite{navon23dws}, namely translate and gaussian noise. The qualitative comparison is presented in Figure~\ref{fig:ssl_qualit}. We show that using our augmentations yields better distinction in the feature space. Additionally, our augmentations achieve a lower NCE loss value of $1.12$ compared to $4.52$ when using the augmentations from ~\cite{navon23dws}.

\begin{figure}
\centering
\begin{minipage}{.45\textwidth}
      \centering
\includegraphics[width=0.9\linewidth]{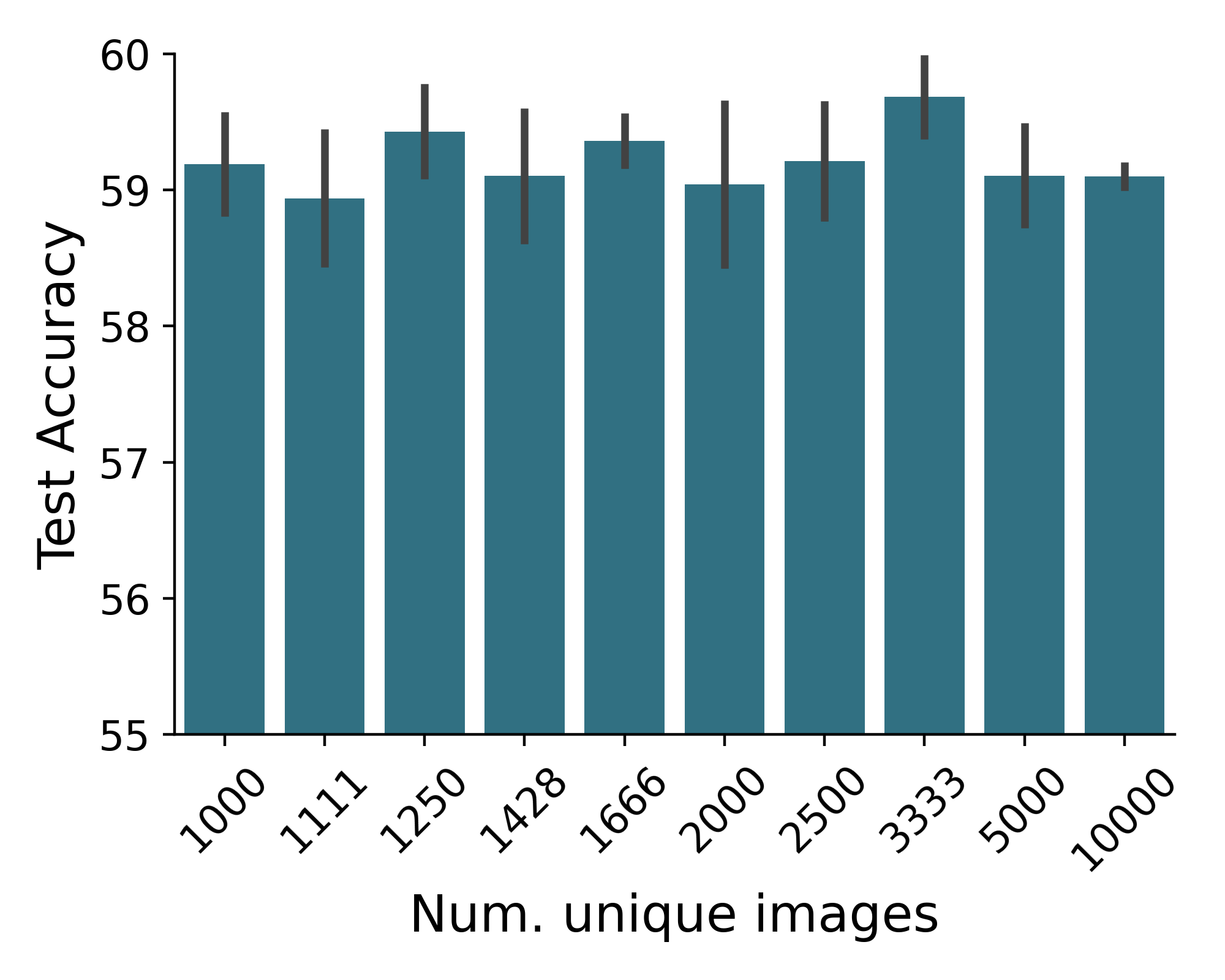}
    \caption{\textit{Additional images or additional neural views}: DWS test accuracy on FMNIST dataset. We fix the number of training INRs and vary the number of unique objects (x axis).}
    \label{fig:view_vs_objects}
\end{minipage}%
\hfill
\begin{minipage}{.45\textwidth}
  \centering
\includegraphics[width=0.8\linewidth]{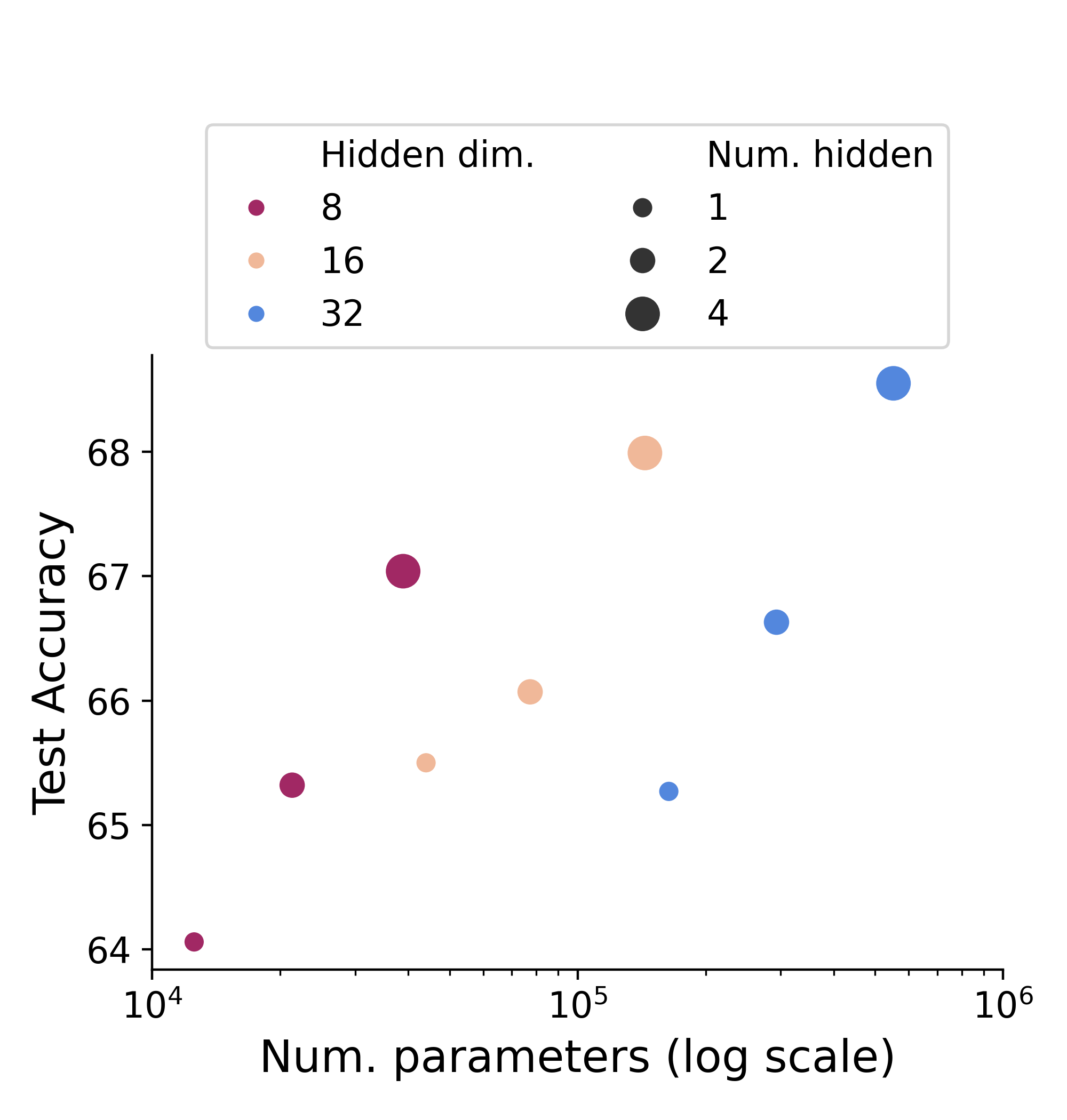}
    \caption{\textit{FMNIST test accuracy as a function of model complexity}}
    \label{fig:dws_capacity}
\end{minipage}
\end{figure}

\subsection{Feature MixUp} \label{app:feature_mixup}
Feature MixUp~\cite{Verma2018ManifoldMB}, also known as Manifold MixUp, is an extended version of MixUp applied to hidden representations. While regular Mixup linearly interpolates between pairs of input samples in the input space, Manifold Mixup operates in a way that considers the non-linear structure of the data manifold. We use the same setup from Section~\cref{sec:inr_classification} and investigate the performance of DWS on the FMNIST dataset using the Feature MixUp augmentation. DWS achieves a test accuracy of $68.47$, which was significantly lower than the MixUp variants presented in this paper.

\newpage
\subsection{MixUp component ablation.} \label{app:mixup_ablation}
\begin{wraptable}[13]{r}{0.5\textwidth}
\vspace{-20pt}
    \centering
\caption{MixUp ablation on FMNIST-INR dataset.}
\label{tab:MixUp_ablation}
    \begin{tabular}{lcc}
    \toprule
 &\multicolumn{2}{c}{FMNIST-INR}\\
        \cmidrule{2-3}
         &  1 View&10 View\\
         \midrule
         No augmentation & $68.30 \pm 0.62$ & $76.01 \pm 1.12$ \\
         \midrule
         Label smoothing&  $69.15 \pm 0.51$&$76.95 \pm 1.74$\\
         Input averaging&  $72.38 \pm 0.39$&$77.92 \pm 0.32$\\
         \midrule
         MixUp&  $74.36 \pm 1.17$&$78.58 \pm 0.20$\\
         Alignment + MixUp&  $\mathbf{75.67} \pm \mathbf{0.36}$ & $\mathbf{79.41} \pm \mathbf{0.56}$\\
         \bottomrule
    \end{tabular}
\end{wraptable}
Weight Space MixUp can be considered as a combination of two basic regularization techniques, one that merges the labels and one that merges the input samples. The first technique is related to label smoothing, a technique where the ground-truth labels are adjusted by perturbations ~\cite{szegedy2016rethinking}. The second technique is defining new inputs by weighted averaging of pairs of samples. Both approaches prevent the model from becoming overly confident and encourage more robust generalization. To investigate the benefit of the individual components, we train DWS on the FMNIST INRs dataset with $3$ different augmentations: (i) label-smoothing, (ii) averaging input data, and (iii) weight space MixUp. Table~\ref{tab:MixUp_ablation} shows that using only label smoothing or input averaging in the weight space helps slightly over no augmentation, but still yields worse results compared to MixUp. It can be concluded from this that the performance boost obtained from weight space MixUp does not result from either of the techniques alone but from the combination of both.





\end{document}